\documentclass[journal=gmj]{CUP-JNL-DTM}%

\usepackage{graphicx}
\usepackage{multicol,multirow}
\usepackage{amsmath,amssymb,amsfonts}
\usepackage{mathrsfs}
\usepackage{amsthm}
\usepackage{rotating}
\usepackage{appendix}
\usepackage{ifpdf}
\usepackage[T1]{fontenc}
\usepackage{newtxtext}
\usepackage{newtxmath}
\usepackage{textcomp}
\usepackage{xcolor}
\usepackage{lipsum}
\usepackage[colorlinks,allcolors=blue]{hyperref}
\usepackage{subfig}

\theoremstyle{definition}

\numberwithin{equation}{section}

\jname{Data/Math}
\articletype{ARTICLE TYPE}
\jyear{YEAR}

\begin{document}

\begin{Frontmatter}

\title[Article Title]{Generative prediction of laser-induced rocket ignition with dynamic latent space representations}

\author[1]{Tony Zahtila}
\author[2]{Ettore Saetta}
\author[1]{Murray Cutforth}
\author[1]{Davy Brouzet}
\author[1]{Diego Rossinelli}
\author[1]{Gianluca Iaccarino}

\authormark{Zahtila \textit{et al}.}

\address[1]{\orgdiv{Center for Turbulence Research}, 
  \orgname{Stanford University}, 
  \country{USA}, \email{tzahtila@stanford.edu}}
  
\address[2]{\orgname{University of Naples Federico II}, 
  \country{Italy}}



\abstract{Accurate and predictive scale-resolving simulations of laser-ignited rocket engines are highly time-consuming because the problem includes turbulent fuel–oxidizer mixing dynamics, laser-induced energy deposition, and high-speed flame growth. This is conflated with the large design space primarily corresponding to the laser operating conditions and target location. To enable rapid exploration and uncertainty quantification, we propose a data-driven surrogate modeling approach that combines convolutional autoencoders (cAEs) with neural ordinary differential equations (neural ODEs). The present target application of an ML-based surrogate model to leading-edge multi-physics turbulence simulation is part of a paradigm shift in the deployment of surrogate models towards increasing real-world complexity. Sequentially, the cAE spatially compresses high-dimensional flow fields into a low-dimensional latent space, wherein the system’s temporal dynamics are learned via neural ODEs. Once trained, the model generates fast spatiotemporal predictions from initial conditions and specified operating inputs. By learning a surrogate to replace the entirety of the time-evolving simulation, the cost of predicting an ignition trial is reduced by several orders of magnitude, allowing efficient exploration of the input parameter space. Further, as the current framework yields a spatiotemporal field prediction, appraisal of the model output's physical grounding is more tractable. This approach marks a significant step toward real-time digital twins for laser-ignited rocket combustors and represents surrogate modeling in a complex system context.}

\end{Frontmatter}

\section*{Impact Statement}
We present a dynamical autoencoder (DnAE) framework that couples cAEs with neural ODEs to predict ignition trajectories in laser-ignited rocket combustors. The problem representation first reduces the simulation data from high-dimensional fields of primitive variables to path-integrated quantities that mimic experimental data acquisition. The resulting quantities are compressed via cAEs to a low-dimensional manifold and subsequently temporal evolution is forecast with parameterized neural ODEs, extending the application of such models beyond existing simple dynamics emulators to a complex engineering problem of interest. The DnAE reduces ignition prediction cost by orders of magnitude relative to large-eddy simulation (LES) while accurately capturing bifurcating ignition success and failure dynamics, representing a step toward real-time digital twins for propulsion systems.


\section{Introduction}
\label{sec:introduction}

Surrogate models are commonly used to accelerate the search for promising designs by replacing expensive evaluations or simulations \citep{forrester2008engineering}, or to estimate variability in quantities of interest (QoI) for an engineering system \citep{sudret2017surrogate}. In many engineering applications, the surrogate represents a mapping $f:\mathbb{R}^d \rightarrow \mathbb{R}$, where the input vector $\boldsymbol{\xi} = (\xi_1,\dots,\xi_d)$ encodes design parameters and variables describing the uncertainty present in the system and the output is a scalar representing local or system-level performance measures. There is no shortage of applications, examples include polynomial chaos expansions (PCE) applied to wake-flow modes \citep{lee2025surrogate} and Gaussian process regression (GPR) used to compare experiments and simulations of a prism bluff-body \citep{duan2019using}. Prediction of full field quantities is more challenging and does not always provide direct engineering insight. Nonetheless, vector-to-vector surrogates of the form $f:\mathbb{R}^d \rightarrow \mathbb{R}^m$ have been explored, where the input encodes parameters or initial conditions and the output is a reduced feature-space representation of the flow field. Examples include parametric extensions of proper orthogonal decomposition (POD) for compressible aerodynamics \citep{bui2003proper}, and temporal extensions to POD {\color{black} such as dynamic mode decomposition (DMD) applied in a predictive capacity to simple partial differential equations \citep{lu2020prediction}}. More recently, for steady fields, non-linear dimensionality reduction has proven to be capable of extreme compression \cite{saetta2024uncertainty} and the flexibility of neural network architectures has facilitated strategies for entire flow-field generation. Subsequent extensions that encompass time-series learning efforts have notably used long short-term memory networks (LSTMs), transformers \citep{solera2024beta} and neural ordinary differential equations (neural ODEs/NODEs) to learn latent spaces of simpler spectrum problems such as the 1D Kuramoto–Sivashinsky equation (KSE) for evolution on an inertial manifold \cite{linot2022data}, or two-dimensional Kolmogorov flow \cite{chakraborty2024divide}.  Neural ODEs provide a simpler, dynamics-consistent framework whose behavior is easier to inspect, in contrast to the comparatively {\color{black} complicated} transformer architectures which are akin to a black-box model.

In the present study, a surrogate framework is applied to a reliability study in the context of rocket ignition success. In particular, we wish to assess binary ignition sensitivities in the combustor, as a function of a high dimensional parameter space. A secondary aim is that the surrogate model provides realistic time-resolved dynamics, which rules out the use of standard supervised classification methods. A natural approach to surrogate modeling is to first build a database, consisting of pre-calculated samples obtained from classical simulation methods, which then train the surrogate model across the parameter vector $\boldsymbol{\xi}$, with components 
$\xi_i$ sampled according to their input probability distributions, e.g.\ 
$\Pr(\xi_i) \sim \mathcal{N}(\mu,\sigma^2)$ or 
$\Pr(\xi_i) \sim \mathcal{U}(a,b)$, and in this study chosen to be experimentally informed where possible \citep{strelau2024laser}. In a laser-ignited rocket combustor context, each time a laser deposits an energy kernel, significant variability in the laser operating conditions \citep{zahtila2023progress} and more generally in the combustor system occurs, meaning that (i) no two ignition trials are the same, and (ii) variations in the system input space $\boldsymbol{\xi}$ can lead to bifurcating outcomes in the system response. Whether a deposited laser-kernel will lead to a kernel evolution sequence with a successful ignition future state depends on the spatiotemporal evolution of the system motivating a vector-to-vector surrogate $f:\mathbb{R}^d \rightarrow \mathbb{R}^m$. To predict whether ignition will occur, among the classical simulation strategies (such as RANS/LES/DNS), a physically accurate and computationally tractable approach is scale-resolving large-eddy simulation (LES), whose predictive ability is validated against experiments \citep{brouzet2025large}. An LES computation of an ignition trial amounts to time-integration of high-dimensional 3-D spatial field and is therefore a costly calculation for the purpose of a single-realization of ignition success. In this work, we carry out an ensemble of LES calculations as ignition trials \citep{zahtila2025bi} to characterize the output space, capturing both input system variability and the resulting multi-modal outcomes.

Building on this ensemble of LES trials, a data matrix of spatiotemporal flow realizations is constructed, $\mathcal{S} = \bigcup_{j=1}^{N} \{ u_j(t_{1}), u_j(t_{2}), \dots, u_j(t_{n}) \}, \quad \text{where } u_j(t_i) \in \mathbb{R}^d$, with $d$ the dimension of the data snapshot, $n$ the number of discrete snapshots per simulation, and $N$ the total number of LES cases. Each individual data snapshot $u_j(t_i)$ comes from a 2-D representation of the 3-D flow-field (detailed later in the manuscript). The complete dataset therefore captures the range of spatiotemporal dynamics across the input uncertainty vector $\boldsymbol{\xi}$ ensemble, and forms the starting point for reduced order modeling. The choice and construction of 2-D representation for the flow-field is crucial, because the present system represents an instance of the compressible reacting Navier-Stokes equations and therefore contains many flow features such as shock–shear layer interactions, vorticity in the shear layer with associated turbulence spectrum, and hydrodynamic ejection caused by laser-induced optical breakdown \citep{wang2020hydrodynamic}. The main idea is firstly, there exists a useful representation that retains the essential physics and with minimal information that we do not target, and secondly, that this low-dimensional representation can be related to the input parameter vector via a learned mapping. We opt for a deterministic autoencoder rather than a variational (VAE) counterpart, to allow the latent space to more clearly capture the bifurcations between ignition success and failure. These two components together are sufficient to build a surrogate model. The present framework therefore nonlinearly spatially compresses the dataset through autoencoders (AEs) and subsequently learns trajectories in the latent space through a neural ODE.

\subsection{Dimensionality reduction}
Dimensionality reduction is a well-established technique in fluid mechanics and has been popular for extraction of {\color{black} flow features and flow modes \citep{brunton2020machine,jaroslawski2025predicting}}. The most popular technique is the proper orthogonal decomposition (POD), $ u(x, t) = \sum_{k=1}^{\infty} a_k(t) \, \phi_k(x)$  where $\phi_k(x)$ are the spatial basis functions (or POD modes) that are orthonormal and capture dominant spatial structures in the data, and $a_k(t)$ are the time-dependent coefficients (modal amplitudes) corresponding to each mode $\phi_k(x)$. POD projects the dataset $\mathcal{S}$ onto an optimal linear basis and this may be useful firstly in inspecting the POD modes to determine the flow structures ranked in terms of energy content, but secondly, a truncated representation can serve as a reduced basis for compression of the original dataset. Although POD is successfully able to identify dominant coherent structures, it often requires a large number of modes to capture multi-scale turbulence \citep{muralidhar2019spatio, alfonsi2007structure} and can become unreliable when strong nonlinear dynamics dominate, motivating alternatives such as DMD \citep{schmid2022dynamic}. One of its machine learning counterparts is the autoencoder (AE), which encodes the data matrix into a compact latent space and subsequently decodes it for reconstruction. For a shallow two-layer AE with linear activation functions, the learned basis recovers the POD modes \citep{milano2002neural}, so that choosing a latent dimension $N_\ell$ corresponds to retaining the first $N_\ell$ POD modes. The latent representation is given by the vector $\mathbf{V} \in \mathbb{R}^{N_\ell}$. {\color{black} When non-linear activation functions are introduced, the network instead learns a non-linear manifold embedded in the full state space $\mathcal{M} \subset \mathbb{R}^m$, allowing more aggressive compression than linear POD \citep{kim2023convolutional,zeng2022data}.}

\subsection{Temporal evolution methods}
After spatial compression, we predict the temporal evolution of the compressed variables: in POD this would correspond to forecasting the modal coefficients $a_k(t)$, while for the autoencoder it corresponds to predicting the dynamics of the latent vector $\mathbf{V}(t)$.

A neural ODE models the continuous evolution of the latent-space variables as a system of ODEs,
\begin{equation}
    \frac{d\mathbf{V}(t)}{dt} = f(\mathbf{V}(t),t; \boldsymbol{\theta}), \label{eq:vanilla_node}
\end{equation}
where $\mathbf{V}(t)$ denotes the latent state at time $t$, and $\boldsymbol{\theta}$ is the learnable parameters of the network. The neural ODE can be considered the continuous extension to ResNet \citep{he2016deep}. It is worthwhile to point out the similarity with the classical semidiscretization procedure, in which the governing equations are discretized in space, thereby reducing the partial differential equations (PDEs) to a system of coupled ODEs in time \citep{ferziger2002computational}. By construction, this NODE formulation is Markovian in the latent space: the future trajectory depends only on the present state $\mathbf{V}(t)$ (and time $t$), and does not explicitly encode long-range temporal memory as in recurrent or attention-based architectures. Once the AE and subsequently NODE architectures are individually trained, a fully deterministic model is constructed, as each physical input vector maps to a unique outcome; this contrasts with previous work on evolution of ignition kernels that applied stochastic strategies based on Wiener processes \citep{chung2024ensemble}.  The NODE framework has for some time been established in its continuous layers design and application \citep{chen2018neural} but typically, usage for time-series prediction has been limited to simpler trajectories such as spirals, or in the case of freely decaying turbulence, input variance through a single parameter, corresponding to initialization of turbulent energy spectrum  \citep{portwood2019turbulence}.

\subsection{State of the art and outline of paper}

In this work, we opt for a machine learning strategy because of the bifurcating dynamics which are highly nonlinear and the need for a framework that can handle parametric dependency as well as state evolution. Linear methods that approach this aim are unsuitable for a variety of reasons.  For instance, there is no expectation to capture bifurcating dynamics through DMD which fits one global linear operator \citep{schmid2010dynamic, lu2020prediction}. The closest linear time-stepping comparison to the present study would be POD spatial compression followed by linear autoregression with exogenous input (ARX) but as noted in preceding work in modeling of a cylinder wake \citep{siegel2008low} a linear model was insufficient and a network architecture was required.  To meet this requirement, we introduce the dynamical autoencoder (DnAE) framework for forecast of laser-induced rocket ignition.

\begin{figure}
    \centering    \includegraphics[width=1.00\linewidth]{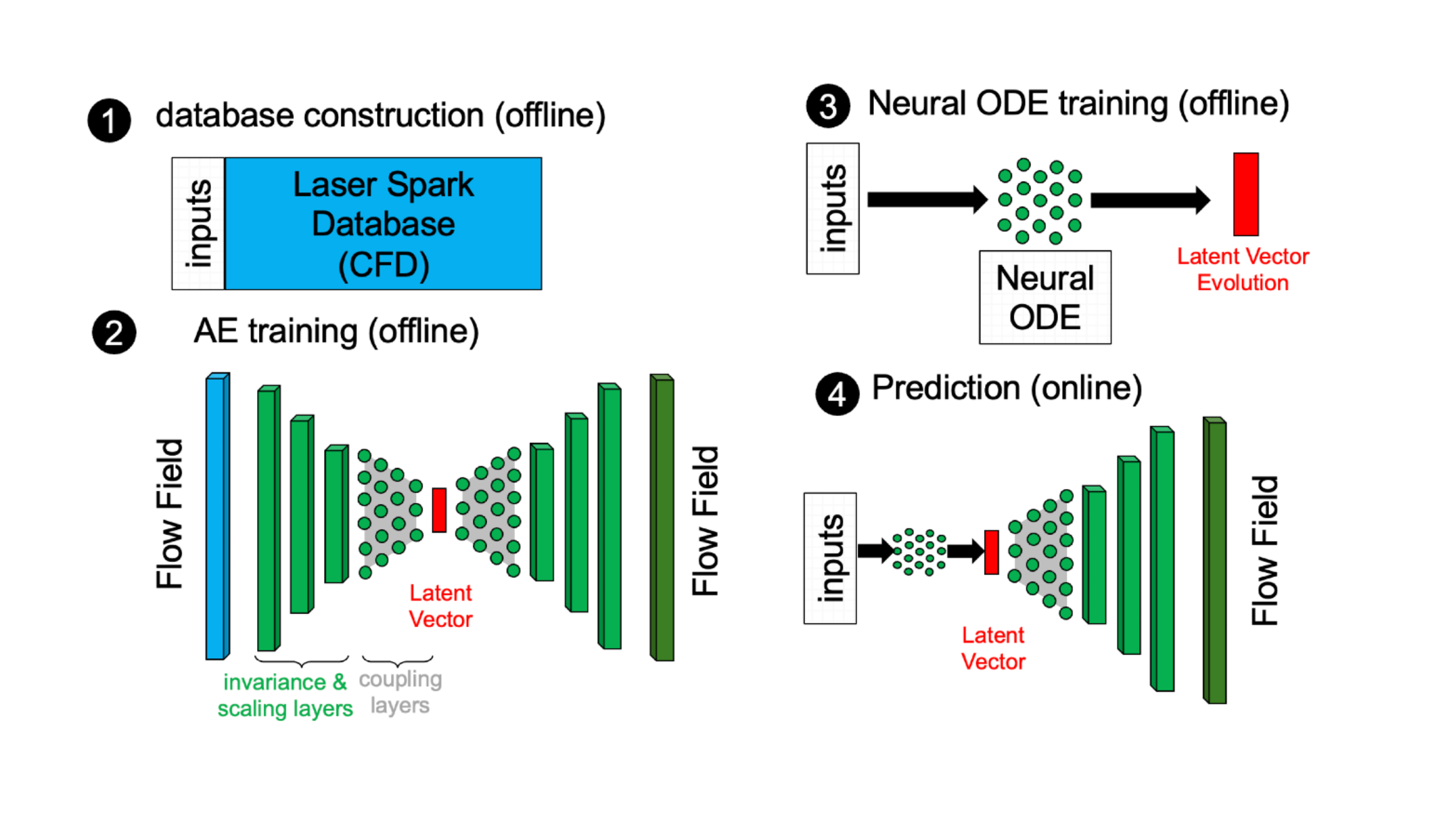}
    \caption{Workflow describing the construction and deployment of the neural ODE and decoder as a generative model.}
    \label{fig:workflow_schematic}
\end{figure}

This paper initially provides a comprehensive description of the input parameter vector in \S \ref{sec:uncertainty_sources}, including the distributional assumptions and parameterization that define system variability. It then outlines the construction and choice of the QoI in \S \ref{sec:ray_tracing}, with possible candidates of several computational path-traced imaging modalities inside the combustor. Following this, the methodology for constructing the surrogate model is detailed in \S \ref{sec:autoencoder} and \S \label{sec:autoencoder}\ref{sec:pnode}, encompassing both spatial dimensionality reduction via autoencoders and the subsequent latent space temporal evolution using neural ODEs. The associated training strategy, including optimization procedures and data handling protocols, is then described in \S \ref{sec:training}. Model performance is then quantitatively assessed in \S \ref{sec:results}. An overall schematic of the data-driven surrogate model construction is given in figure \ref{fig:workflow_schematic}. With the surrogate model therefore ready for deployment, generative trajectories are obtained for number of ignition trials $N_t = 10^{6}$ and an enhanced understanding of the ignition response to the input space, which enables decision boundaries for successful rocket ignition greatly improving upon the sparse and under-converged ignition probability maps obtained with the limited LES simulations available, presented in \S \ref{sec:generative_sec}. 

\section{Data and Methodology}
\label{sec:methodology_data}

\subsection{Sources of uncertainty and data representation}
\label{sec:uncertainty_sources}

The ensemble of LES ignition trials described in §1 provides a data-driven view of how different operating conditions lead to ignition success or failure. To interpret and generalize these outcomes, we formalize the system variability within a structured uncertainty quantification (UQ) campaign \citep{cutforth2025bi}. This framework identifies fifteen input parameters (summarized in Table~\ref{tab:uncertainties} in appendix \ref{app:uncertainty}) that encode variability (i.e. uncertainty) originating from the companion experimental campaign, the combustor system, and model-form assumptions. These inputs form the parametric drivers for the surrogate model. Briefly, the uncertainty set includes epistemic variations in laser spark characteristics and focal accuracy ($\xi_0$–$\xi_5$). Deposition timing is influenced by two distinct mechanisms: (a) small-scale lags that introduce aleatoric uncertainty through turbulence-driven shifts ($\xi_6$), and (b) large-scale lags that capture epistemic uncertainty due to variability in the available methane system mass ($\xi_7$). Additional sources include modeling uncertainties in flame propagation ($\xi_{8}$–$\xi_{10}$), model-form uncertainty in the sub-grid stress coefficient ($\xi_{11}$), epistemic uncertainties tied to fuel–oxidizer intake conditions ($\xi_{12}$–$\xi_{13}$), and geometric variability in combustor cross-section ($\xi_{14}$).

\begin{figure}
    \centering
    \includegraphics[width=0.99\linewidth]{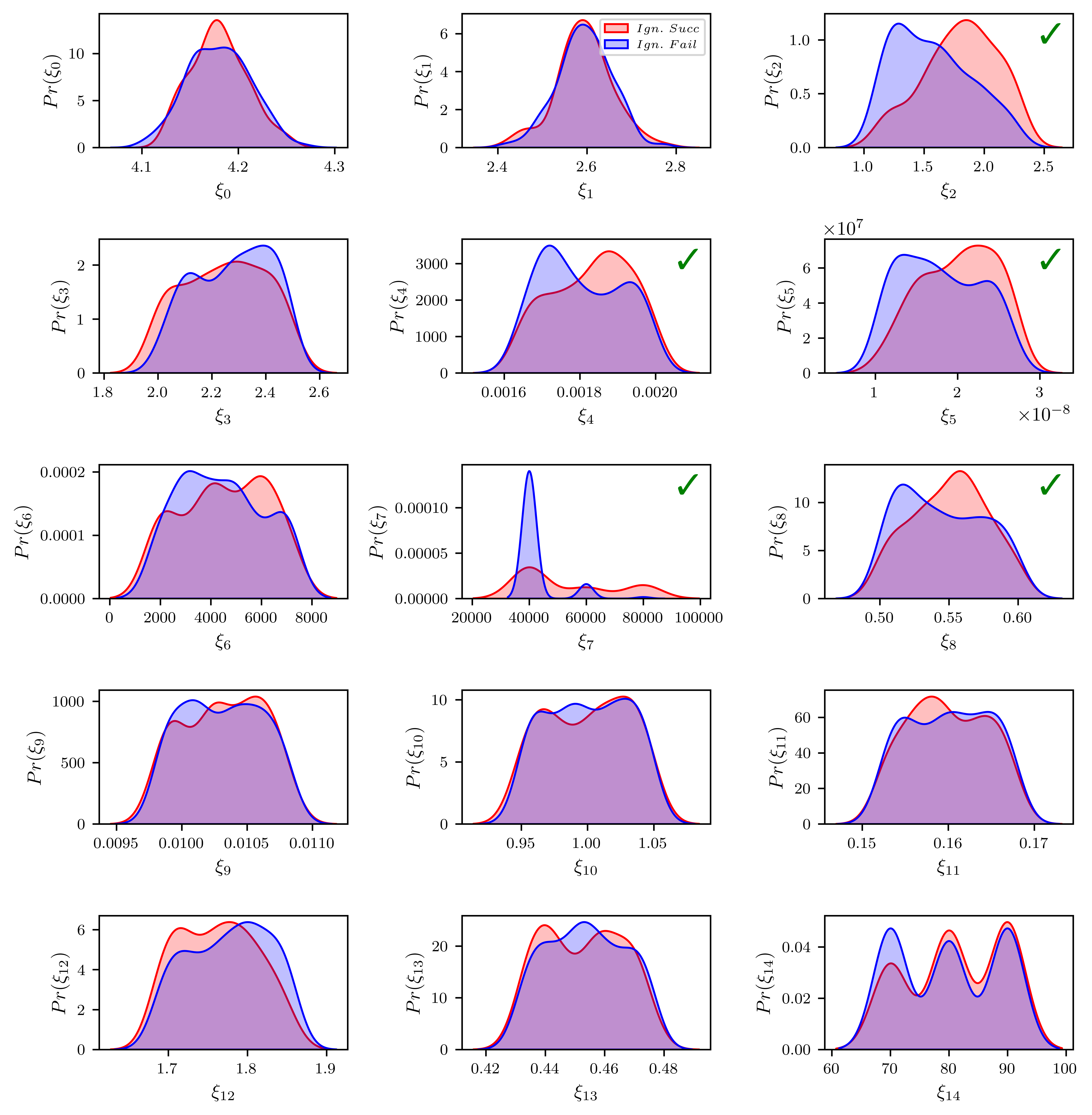}
    \caption{Ignition probability density functions (PDFs) marginalized over the input parameters $\xi_0$ through $\xi_{14}$, 
    comparing ignition success (red) and ignition failure (blue) through kernel density estimates; green check marks denote the five parameters with the largest KL divergence between igniting and non-igniting cases. Full description of parameters in Table \ref{tab:uncertainties} in Appendix \ref{app:uncertainty}.}
    \label{fig:distribution_figure}
\end{figure}

A central aim of the UQ campaign was to assess ignition sensitivities in the combustor and each of the LES computations in the ensemble of ignition trials leads to a response of ignition failure or success. To this end, Figure~\ref{fig:distribution_figure} shows marginal distributions of the input parameters, separated into igniting and non-igniting cases. While some distributions differ only slightly (reflecting sampling convergence), others exhibit clear structural differences. To quantify these differences, the Kullback–Leibler (KL) divergence \citep{kullback1951information} was computed between the success and failure groups, with parameters exhibiting the largest divergence highlighted in Figure~\ref{fig:distribution_figure} with green check marks. However, ignition is fundamentally a Bernoulli random variable (success/failure), even if a continuous proxy is used such as the maximum chamber pressure over time instead of a Boolean label, the underlying problem of predicting a sharp regime transition between ignition and failure is not removed. Consequently, estimating reliable probabilities is inherently slow to converge, e.g., with $N = 100$ independent trials, the Wilson score interval \citep{wilson1927probable}, which provides a confidence range for a binomial proportion accounting for finite sample size, for a nominal $p = 0.5$ ignition probability still spans roughly $[0.40, 0.60]$, highlighting substantial uncertainty. Given finite compute allocations and therefore limited sampling, the available LES ensemble is insufficient for robust convergence, and this motivates the construction of the DnAE surrogate model to efficiently augment ignition probability quantification. A comprehensive joint characterization of the most informative parameters is deferred; we return to it later using surrogate-enabled Monte Carlo.

To assemble a statistically useful LES dataset for reduced order modeling, we generated 300 ignition trials by first constructing tailored LES \cite{cutforth2025bi} built from large-scale experimentally validated calculations \cite{brouzet2025large, passiatore2024computational}. The tailored LES computations differ from the increased scale-resolving calculations (for which $\mathcal{O}(1\text{--}10)$ are feasible) because they employ a significantly coarser mesh and feature reduced chemistry, and are calibrated by employing an Ensemble Kalman filter (EnKF) for deterministic processes. This procedure matched the tailored LES to emulate the high-fidelity laser kernel dynamics, and turbulence quantities in the co-flowing jet. With this calibration in place, larger ensembles became feasible.  The resulting dataset enabled construction of ignition probability maps, where ignition success was treated as a {\color{black} Bernoulli random variable  ($Y \in \{0,1\}$)}. Probabilities regressed onto a logistic curve as a function of radial focal position were consistent with experimental probability maps both in terms of the mean predicted ignition probability and the $95\%$ confidence interval indicating a computational ensemble that was statistically consistent with the laboratory model rocket-combustor experiments \citep{strelau2024laser}.


\subsection{Quantity of interest: Thermal imaging}
\label{sec:ray_tracing}

\begin{figure}
    \centering
    \includegraphics[width=0.99\linewidth]{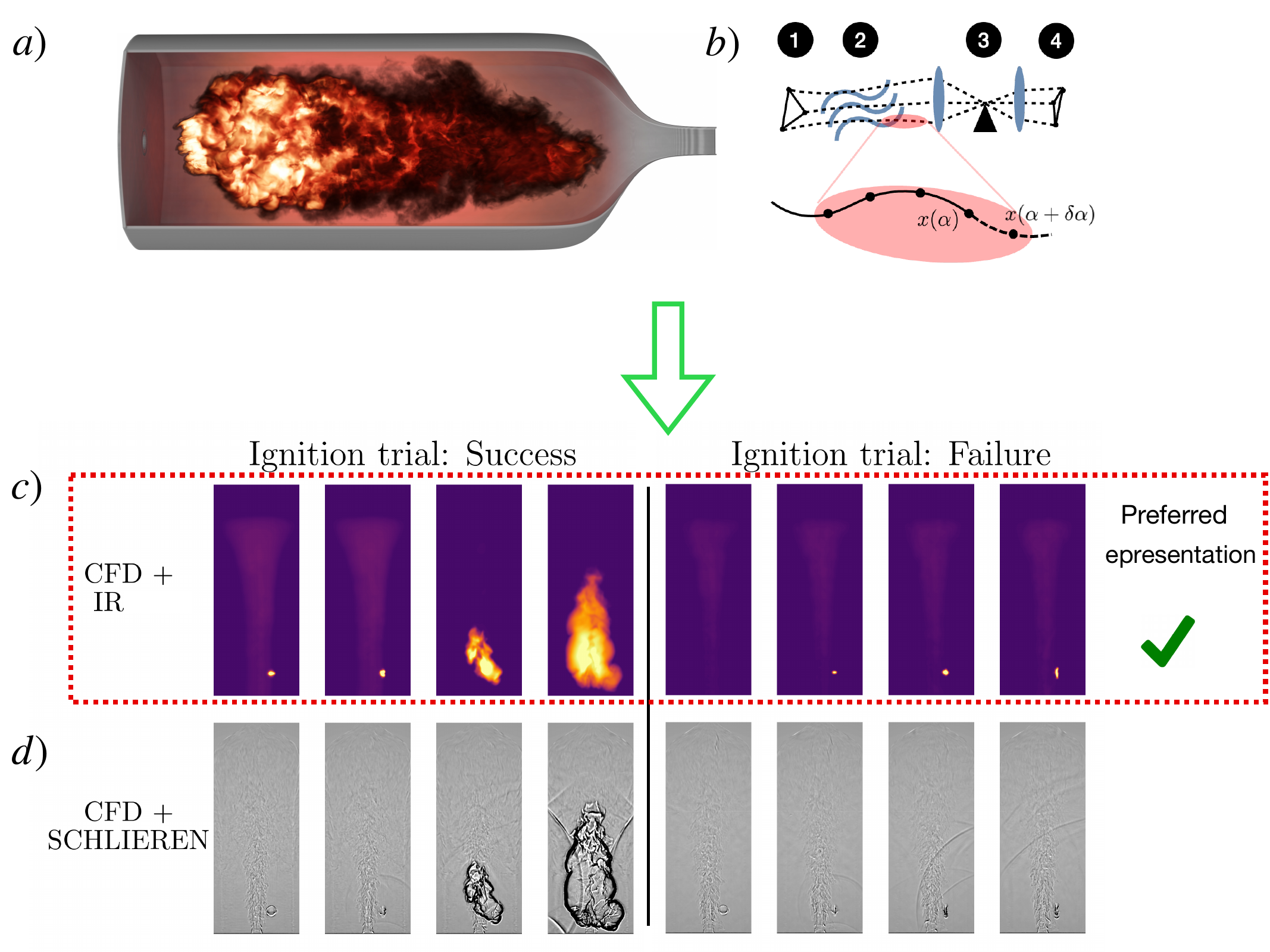}
    \caption{Quantity of interest: spatiotemporal fields, (a) rendered simulation output of the successfully ignited combustor, (b) illustration of the integrated quantity acquisition, (1) light beam generation, (2) ray evolution according to the {\color{black} Eikonal equation}, (3) Snell's law and occlusion, and (4) irradiance sampling, (c) successful and unsuccessful time-series for the computational infrared, and (d) computational Schlieren.}
    \label{fig:qoi_figure}
\end{figure}

Having generated a large and statistically useful tailored LES data ensemble, we proceed to select an appropriate representation of the combustor dynamics on which to train and develop the surrogate model. In this predictive rocket-ignition context, we therefore require a reduced data representation that remains physically meaningful yet tractable for learning. A further requirement for validation would be that the representation is relatable to the observables from experimental sensing modalities. Machine learning surrogate models trained on reduced or structure-promoting representations are preferable in turbulence due to the extreme high dimensionality, chaotic nature, and multiscale features of full 3-D fields. Recent examples include, in the context of multiphase flows, investigation on the choice of interface treatment through diffuse, sharp, and level set approaches \citep{cutforth2025convolutional}, and in urban street canyons, 2-D particle-image-velocimetry (PIV) was preferable at a judiciously chosen plane that captured turbulence quadrant events \citep{jaroslawski2025predicting}. These reduced forms capture essential dynamics while significantly lowering computational cost and mitigating overfitting, making learning more tractable and generalizable. The essential features in the combustor include (1) co-flowing (fuel and oxidizer) turbulent intake, (2) deposited energy kernel location and dynamics, and in the successful ignition scenario, (3) flame propagation and stable combustion. 



Figure \ref{fig:qoi_figure}(a) displays rendered output from the CFD solver \citep{di2020htr} and the associated flame visualization.  
{\color{black}
To emulate the experimental data acquisition, our first approach is to computationally emulate high-speed Schlieren by integrating light beams for a collimated light source passing through the variable-density flow. The trajectories obey the governing equation derived from Fermat's principle:

\begin{equation}
    \frac{\partial}{\partial \alpha}\left(n \frac{\partial \mathbf{x}}{\partial \alpha} \right) = \nabla n
\end{equation}

\noindent
where $\mathbf{x}(\alpha)$ is the light beam trajectory as a function of arc length $\alpha$, $n$ is the refractive index field calculated from the species mass fractions and Gladstone-Dale mixture relation, with full details in supplementary material of \cite{brouzet2025thermal}. {\color{black} The path-tracing approach illustrated in Figure \ref{fig:qoi_figure}(b) mimics experimental line-of-sight integration to reduce the data to a 2D signal.} While informative, this high-dimensional representation includes numerous flow and combustion features that are not necessarily predictive of ignition success.

The computational infrared (IR) imaging emulates the methodology used for chemiluminescence modeling \cite{brouzet2025thermal} and instead of integrating refractive beam trajectories, this method relies on the orthographic projection of a volumetric emission field. While the chemiluminescence model utilizes species concentrations to define the source term, radiative emission is modeled to reproduce thermal imaging from the temperature field.
}
 Representative time-series for both successful and failed ignition events are shown using IR imagery in Figure \ref{fig:qoi_figure}(c) and computational Schlieren imaging in Figure \ref{fig:qoi_figure}(d). 
 Although the Schlieren images capture a wide range of turbulent spatial scales, we prioritized the infrared data due to its more informative representation of spark location and subsequent dynamical evolution of the laser-deposited energy kernel. We conducted $N_c = 300$ CFD cases with $N_s = 20$ aligned post-deposition snapshots. 
 This resolution captured the essential kernel dynamics and combustion evolution while balancing memory requirements to fit the resultant data matrix on a single GPU.

\subsection{Autoencoder}
\label{sec:autoencoder}

\begin{figure}
    \centering
    \includegraphics[width=0.89\linewidth]{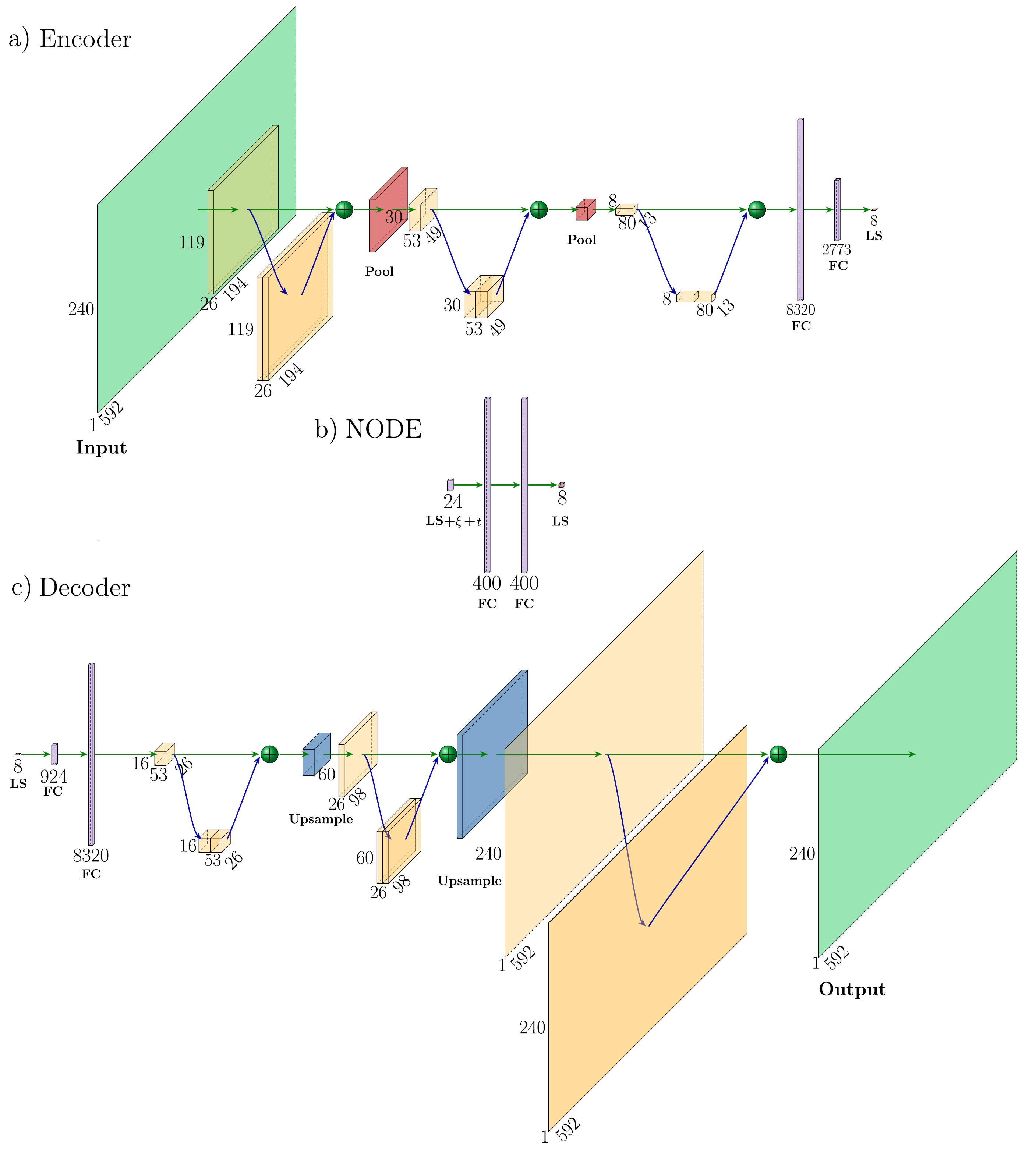}
    \caption{Architecture schematic. (a) Encoder: a sequence of convolutional, with residual connections and pooling blocks maps the single–channel input to progressively smaller but deeper feature maps, which are then flattened and passed through fully-connected layers to obtain an 8-dimensional latent state $\mathbf{V}$. (b) NODE: the latent state is augmented with the control inputs ($\xi,t$), with two fully connected layers and the latent vector  $\mathbf{V}$ is forecast for the duration of the ignition trial. (c) Decoder: the latent state $\mathbf{V}$ is lifted back to high dimension by fully-connected layers and a sequence of convolutional and upsampling blocks reconstructing the output field.}
    \label{fig:DnAE}
\end{figure}

Despite the spatial reduction to 2D, the IR fields described in \S2.2 must be reduced in dimensionality in order to enable temporal modeling in the first place. We choose nonlinear cAEs because they have been demonstrated to yield efficient compression while retaining essential physical structures \citep{saetta2024uncertainty}, however, alternative approaches for learning latent manifolds are also available \citep{fan2025physically}. In the present framework, we therefore employ a cAE that learns a low-dimensional latent representation using an architecture composed of three convolutional blocks that progressively extract multi-scale spatial features while reducing resolution. These blocks utilize increasing channel counts and intervening pooling operations, augmented with residual connections and nonlinear activation functions (ReLU). Subsequently, fully connected layers nonlinearly combine these multiresolution features to compress the $\mathbf{x} \in \mathbb{R}^{592 \times 240}$ input IR fields to an 8-dimensional latent state. Finally, the decoder mirrors this configuration, employing symmetric upsampling and convolutional blocks to reconstruct the full high-resolution input space from $\mathbf{V}$ (see Figure~\ref{fig:DnAE}a,c).

%

As a preprocessing step, each image matrix is min–max normalized to the range $[0,1]$ through affine rescaling, ensuring consistent numerical scale and improved conditioning during training \cite{LeCun2012}. 
Each normalized light–intensity field is passed to the AE as a single-channel 2D input, on which the network performs a nonlinear encoding–decoding operation:
\begin{equation}
\mathbf{V} = e(\mathbf{x}), \quad 
\hat{\mathbf{x}} = d(\mathbf{V}), \quad 
\hat{\mathbf{x}} = d(e(\mathbf{x})),
\end{equation}
where $e: \mathbb{R}^{m \times n} \rightarrow \mathbb{R}^{N_\ell}$ is the encoder mapping for an input with dimensions $m \times n$, and $d: \mathbb{R}^{N_\ell} \rightarrow \mathbb{R}^{m \times n}$ the decoder mapping. Here $\mathbf{x}$ denotes the input field, $\hat{\mathbf{x}}$ its reconstruction, and $\mathbf{V}\in\mathbb{R}^{N_\ell}$ the latent variables. Following \citep{saetta2022machine}, the optimization objective is a regularized reconstruction loss,  
\begin{equation}
\mathcal{L} = \frac{1}{N}\sum_{i=1}^N \lVert \mathbf{x}_i - \hat{\mathbf{x}}_i \rVert_2^2 
+ \beta \, \frac{1}{N_e} \sum_{j=1}^{N_e} (l^e_j)^2 ,
\end{equation}
\noindent

where $l^e_j$ are the encoder weights, $N_e$ is their total number, and $\beta$ is a user-defined regularization coefficient. In our implementation $\beta = 10^{-6}$ with $L_2$ regularization, consistent with prior work on aerodynamic autoencoders.  The latent space dimension $N_\ell$ is a critical hyperparameter which controls the latent space manifold that captures essential flow physics.  

A key property of this construction is that the autoencoder is deterministic: for any unseen input $\mathbf{x}$, the encoder yields a unique latent vector $\mathbf{V}=e(\mathbf{x})$, which the decoder deterministically maps back to the reconstructed field $\hat{\mathbf{x}}=d(\mathbf{V})$. In contrast to variational autoencoder (VAE) approaches {\color{black} that introduce stochasticity} in the latent space \citep{fan2025physically}, this deterministic structure ensures reproducibility of trajectories and avoids additional assumptions on the latent distribution \citep{saetta2024uncertainty}. Once the latent space is established, the decoder can be deployed as a generative model: new realizations can be generated as $\mathbf{V}(\boldsymbol{\xi}) = \mathcal{I}(\{\mathbf{V}_i\},\boldsymbol{\xi})$, where $\mathcal{I}$ denotes an interpolation operator acting on the latent ensemble $\{\mathbf{V}_i\}$ with associated input parameter $\boldsymbol{\xi}$, yielding $\hat{\mathbf{x}}(\boldsymbol{\xi}) = d(\mathbf{V}(\boldsymbol{\xi}))$.  This interpolation provides a continuous parametrization of the solution manifold, enabling exploration of unseen operating conditions. In this sense, the AE not only compresses but also embeds in the reduced space a generative capability, which we exploit in the subsequent temporal modeling of ignition trajectories.

\subsection{Parameterized neural ODE}
\label{sec:pnode}

The temporal dynamics of the latent-space representation $\mathbf{V}(t)$ are advanced using a parameterized NODE framework to model the continuous evolution of the latent-space parameters, which forms a system of coupled ODEs,
\begin{equation}
    \frac{d\mathbf{V}(t)}{dt} = f(\boldsymbol{\xi}, t,\mathbf{V}(t); \boldsymbol{\theta}), \label{eq:param_node}
\end{equation}
where the present model form differs from the vanilla neural ODE shown in \eqref{eq:vanilla_node} by incorporation of the input parameter vector $\boldsymbol{\xi}$, and as a reminder $\boldsymbol{\theta}$ is the learnable parameters of the network.  The vanilla neural ODE is sufficient, for example, to emulate the latent space dynamics for flow past a cylinder with a single instance of the parameter space \citep{rojas2021reduced} but the parameterized extension accounts for variations in the input parameters \citep{lee2021parameterized}. Instead of {\color{black} introducing uncertainty via a VAE} in the preceding dimensionality reduction stage stage, we handle parametric uncertainty here by conditioning the latent NODE dynamics directly on the input vector $\boldsymbol{\xi}$.

The network in this study is a simple feed-forward network but additional network features can be added such as through convolutional layers \citep{shankar2020learning}. Here, $f(\cdot)$ is a multilayer perceptron (MLP) with {\color{black} two hidden layers of width $w$} each with $\tanh$ activation, shown within the DnAE in Figure~\ref{fig:DnAE}(b), mapping the input state $(t,\mathbf{V}(t),\boldsymbol{\xi})$ to the latent derivatives $\dot{\mathbf{V}}(t)$. Formally, the latent trajectory can be written as an initial value problem \citep{chen2018neural}:
\begin{equation}
    \mathbf{V}(t) = \mathbf{V}(t_0) + \int_{t_0}^{t} f\!\left(\tau, \mathbf{V}(\tau), \boldsymbol{\xi}; \boldsymbol{\theta}\right)\, d\tau,
\end{equation}
which expresses the time-series as the initial condition plus the accumulated effect of the latent dynamics. 
In the current study, this integral is approximated using a fixed-step fourth-order Runge--Kutta scheme. 
Therefore, given an initial latent state $\mathbf{V}(t_0)$ from the AE compression, the trajectory $\hat{\mathbf{V}}(t)$ is propagated over discrete times $\{t_i\}_{i=0}^T$ by
\begin{equation}
    \hat{\mathbf{V}}(t_{i+1}) = \hat{\mathbf{V}}(t_i) + \text{RK4}\!\left(f, t_i, \Delta t, \hat{\mathbf{V}}(t_i), \boldsymbol{\xi}\right),
\end{equation}
where $\Delta t = t_{i+1} - t_i$.

The loss term was specified to enforce trajectory reconstruction accuracy and consistency of temporal derivatives. Specifically,
\begin{equation}
\mathcal{L} = \frac{1}{NT} \sum_{i=1}^N \sum_{j=1}^T \big\| \mathbf{V}_j^{(i)} - \hat{\mathbf{V}}_j^{(i)} \big\|_2^2 
+ \lambda \, \frac{1}{N(T-2)} \sum_{i=1}^N \sum_{j=2}^{T-1} \big\| \dot{\mathbf{V}}_j^{(i)} - \dot{\hat{\mathbf{V}}}_j^{(i)} \big\|_2^2,
\label{eq:loss_func}
\end{equation}
where $\mathbf{V}_j^{(i)}$ is the true latent trajectory, $\hat{\mathbf{V}}_j^{(i)}$ the NODE prediction, and $\dot{\mathbf{V}}_j^{(i)}$ is a second-order finite-difference approximation of the velocity of the latent space vector, and the weighting hyperparameter $\lambda = 10^{-2}$ balances the overall contribution of the curvature penalty which is the last term in \ref{eq:loss_func}. The inclusion of the curvature penalty is here a soft constraint to encourage the neural ODE to learn the directional changes, such as in the latent space bifurcation that occurs in ignition.

Briefly we remark on some present limitations. Separating time and space in our surrogate model's prediction for evolving temperature fields introduces limitations in their physical consistency. As evolution is handled fully by the neural ODE in a low-dimensional latent state, spatial interactions become effectively non-local through the learned autoencoder representation, weakening the system causality. Secondly, because the latent space itself is not tied to physical conserved quantities, it is difficult to enforce constraints such as conservation laws directly; physics-informed NODE variants such as PINODE \citep{sholokhov2023physics} partially address this by embedding PDE structure into the dynamics, but they require explicit knowledge of the governing equations which we do not have for our IR fields.

\section{Training}
\label{sec:training}

\begin{figure}
    \centering
    \includegraphics[width=1.0\linewidth]{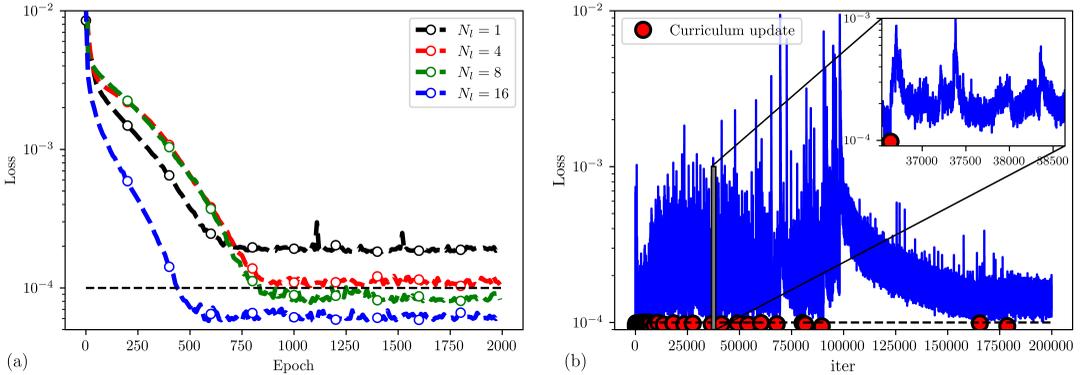}
    \caption{(a) Evolution of the validation loss against epoch during the training of the AE, for choices of latent-space dimension hyperparameter  $N_\ell = 1, 4, 8$ and $16$.   (b) Representative loss during training of the NODE for the 200{,}000 iterations, for choice of hidden layer width hyperparameter $w = 400$. The horizontal black line indicates the loss criterion for a curriculum learning increment,  and markers at spike locations denote curriculum updates.}
    \label{fig:network_training}
\end{figure}

For the bifurcating physics and complex spectrum in rocket ignition and combustion, the effectiveness of the proposed framework relies critically on training methodology; we therefore describe the procedures adopted for the AE and neural ODE to robustly learn the compressed spatial fields and latent trajectories. The spatial and temporal components are trained separately to simplify the schedule and more importantly, joint optimization incurs higher computational cost without necessarily yielding performance gains \citep{vlachas2022multiscale}. For spatial compression, the encoder reduces the input data matrix $\mathcal{S}$ to a compact latent space, and the decoder mirrors this architecture for reconstruction. Confidence in the network design stems from prior work on spatial compression of flow fields \citep{saetta2022machine} and preliminary studies therein evaluating the influence of architectural hyperparameters (e.g., number of layers, activation functions). The key latent space dimension, $N_l$, is varied, and corresponding training results are presented in Figure~\ref{fig:network_training}(a). The validation performance for $N_l = 4, 8,$ and $16$ shows a continuous, albeit small, improvement. We selected $N_l = 8$ as a trade-off, offering good compression while retaining a relatively low latent-space dimensionality. As an aside, when the computational Schlieren in Figure \ref{fig:qoi_figure}(d) is used for training, the scale-retaining features of this imaging modality required a significantly higher latent space dimension before meaningful compression was achieved.

Training NODEs on complex trajectories requires backpropagating through recursively applied deep neural layers. Despite standard stabilizations, this integration depth often leads to exploding or vanishing gradients and reduced sensitivity \citep{chakraborty2024divide}. Several training strategies have been designed to overcome the reduced sensitivity for neural ODEs, such as shooting methods \citep{turan2021multiple} to capture higher frequency responses, and another strategy was the multistep penalty method \citep{chakraborty2024divide} that split trajectories into multiple, non-overlapping time windows and included an optimization loss term to penalize the window discontinuities. In the present work, vanilla training of the parameterized NODE model resulted in a loss function that failed to decay, and the learned solution trajectories merely suppressed oscillatory components rather than capturing the true underlying dynamics. To address this, a curriculum learning strategy was introduced, which can be considered a continuous analogue of the multistep penalty method. In general, curriculum learning structures the optimization process such that training begins with simpler tasks and gradually progresses to more complex ones, thereby improving both convergence and generalization \citep{bucci2023curriculum}. 

In the present application, task difficulty was defined by the temporal extent of the latent-space trajectories. To clarify, all training samples of the parameter space $\boldsymbol{\xi}$ were available for learning but only a restricted time interval of each trajectory. This contrasts the curriculum regularization approach \citep{krishnapriyan2021characterizing} used in training of physics informed neural networks (PINNs) that progressively samples different parts of the parameter space leading to lower error rates. Specifically, let $\{t_j\}_{j=1}^T$ denote the discrete time indices of the latent trajectory $V(t)$. Curriculum learning was implemented as a staged optimization:

\begin{align}
    & \text{Stage loss:} \quad 
    \mathcal{L}(T) = \frac{1}{T}\sum_{j=1}^{T} \big\| \mathbf{V}(t_j) - \hat{\mathbf{V}}(t_j) \big\|^2, \label{eq:curr_loss} \\[6pt]
    & \text{Update rule:} \quad 
    T \;\mapsto\; T + \Delta T 
    \quad \text{whenever } \mathcal{L}(T) < \epsilon. \label{eq:curr_update}
\end{align}

\noindent
where the full time-series of the trajectory ensemble was divided into $N_L = 50$ folds, chosen as the maximum number that partitions the signal into integer-length segments. The curriculum update tolerance was set to $\epsilon = 1 \times 10^{-4}$, which by inspection corresponded to accurate capture of the characteristic curvature of the trajectories.

\begin{figure}
    \centering
    \includegraphics[width=1.0\linewidth]{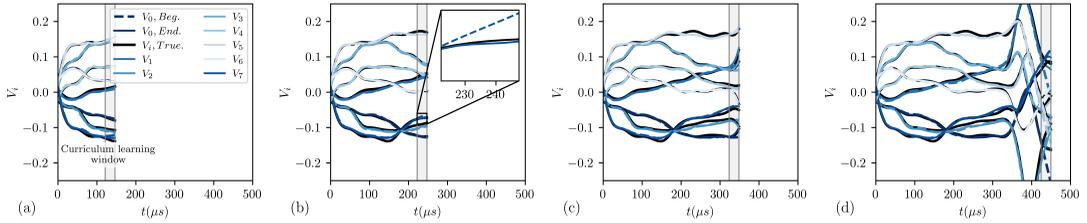}
    \caption{Representative case of latent-space trajectories ${V}_i$ illustrating the curriculum learning process. Panels (a)–(d) display the advancing training window (gray band), highlighting the progressive learning of newly revealed trajectory segments as the time horizon extends. Dashed lines indicate the beginning prediction at the start of the update, while solid lines show the final learned trajectory.}
    \label{fig:curriculim_learning_illustration}
\end{figure}

A drawback to the curriculum training method in \eqref{eq:curr_loss}-\eqref{eq:curr_update} as compared with the discontinuous multistep penalty method is there is no guarantee that the previously learned trajectory will retain its level of accuracy on the preceding segment, but in practice this was not an issue as the loss function retains contribution from the entire segment being learned. Figure~\ref{fig:network_training}(b) shows the loss function for the neural ODE, with curriculum update events marked. Note the inset of Figure~\ref{fig:network_training}(b) shows that after the curriculum is extended, the loss initially spikes as there is more signal to learn, and the loss thereafter decays. The model is trained with the Adam optimizer and the learning rate was initialized at $lr = 10^{-3}$, reduced by an order of magnitude after 100{,}000 iterations, and decreased again at 150{,}000 iterations. Figure~\ref{fig:curriculim_learning_illustration} illustrates the curriculum learning process for the latent-space components $\mathbf{V}$ of a representative trajectory. The NODE trajectory immediately following a curriculum update is shown alongside the final learned trajectory. As highlighted in the inset of Figure~\ref{fig:curriculim_learning_illustration}(b), the initial NODE prediction diverges in the curriculum learning window from the ground-truth trajectory. This region is prioritized by the gradient-based optimization process because it contributes the largest error relative to the previously learned segment.

The present training strategy uses a single NVIDIA Tesla P100 (16 GB) GPU for 24 hours to efficiently train the AE to spatially compress the data matrix $\mathcal{S}$ into a learnable latent space and subsequently, training the parameterized NODE takes 48 hours on a CPU. The key to the present complex problem becoming learning in a realistic time-frame was (i) a problem representation that retained the laser-spark kernel evolution dynamics, and combustion features, but is not distracted by the multi-scale complex spectrum of turbulence present in the numerical schlieren of a co-flowing jet, and (ii) a curriculum learning strategy that allowed for complex trajectories with multiple modes and bifurcations to be learned.

\section{Results}
\label{sec:results}

\subsection{Latent space trajectories}
\label{sec:latent_space_ae_trajects}

\begin{figure}
    \centering
    \includegraphics[width=1.0\linewidth]{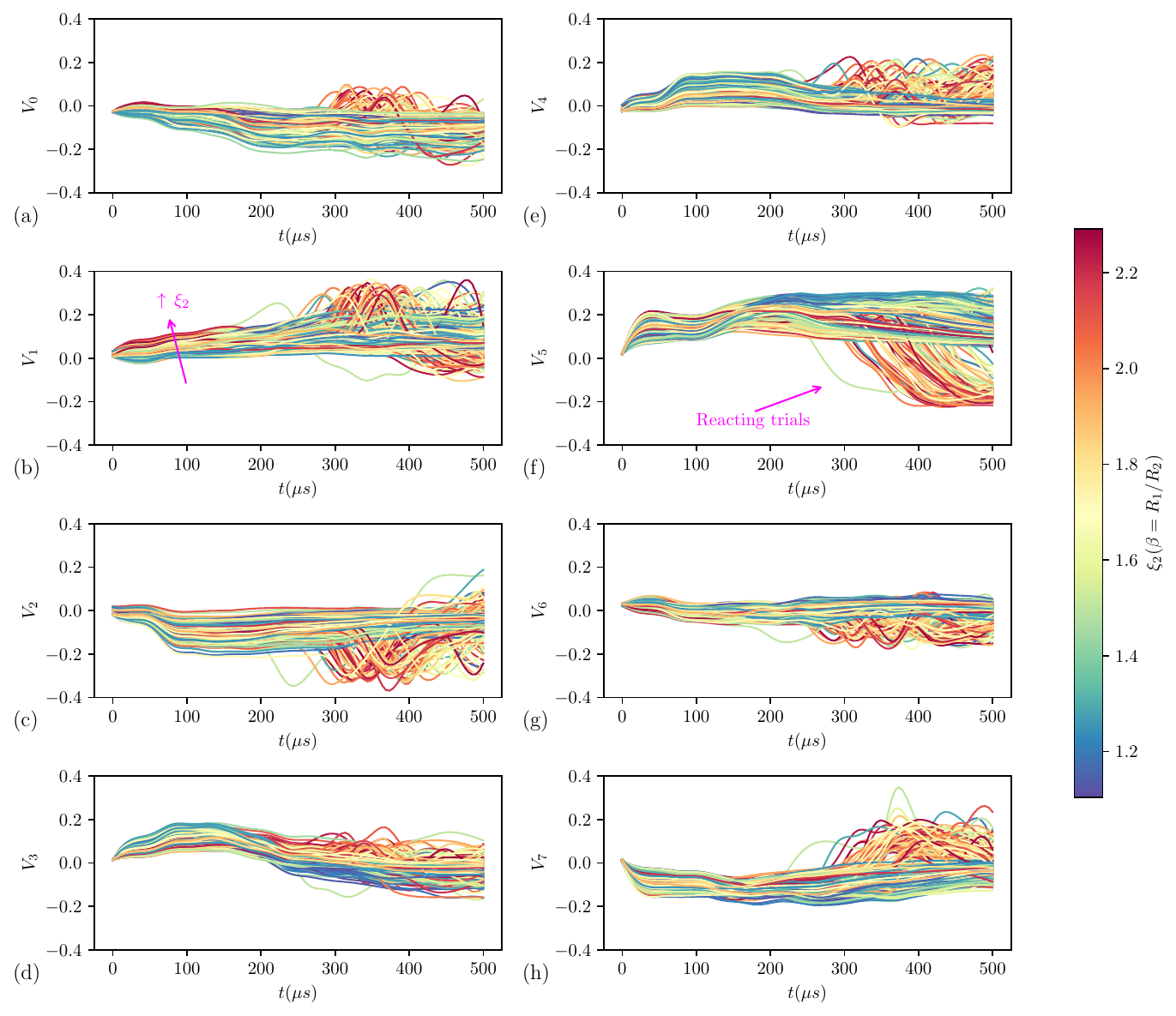}
    \caption{Latent variables, up-sampled in time, for the reacting rocket combustor LES ensemble, post-processed with coloring by a component of the input vector, $\xi_2 = \beta= R_1/R_2$.}
    \label{fig:autoencoder_latent_space}
\end{figure}

To assess whether the trained model captures the essential dynamics of ignition, we first examine the latent-space trajectories, i.e., the mapping of the IR sequences from image space $\mathbf{x}(t_j) \in \mathbb{R}^{m \times n}$ ($m, n= 592 \times 240$) by the AE into low-dimensional vectors $\mathbf{V} \in \mathbb{R}^{N_\ell}$ that is required to extract meaningful latent representations. At the time of spatial compression, the AE is not temporally aware and simply encodes possible combustor states; as a post-processing step for analysis, we order the latent vectors in time $\mathbf{V}(t_j)$ for each trial with unique sample of $\boldsymbol{\xi}$, yielding a family of trajectories. Figure~\ref{fig:autoencoder_latent_space} displays each latent component $V_i$ as a time-series, which after linear up-sampling form smooth curves. At early times, the coordinates capture the initial stages of laser deposition and kernel formation, with variance across cases small and tightly clustered. Subsequently, the trajectories reflect interaction with the co-flowing fuel–oxidizer shear layer, and finally, when $t \gtrsim 250 \,\mu\text{s}$ the bifurcating outcomes of either: (i) runaway combustion/ignition success, or (ii) a dissipating kernel/ignition failure.

The latent space thus provides a reduced representation that preserves essential dynamics in compact form. For reference, in previous work on time-averaged aerodynamics, systematic analysis of latent components yielded physical insight into inviscid and viscous contributions \cite{saetta4898782autoencoders,saetta2024uncertainty}. This was achieved by sequentially training and conditioning the latent space to separate physical processes. In the present application, the latent space could therefore be interrogated by systematically deactivating components of the latent space vector $\mathbf{V}$ to uncover what physics each component $V_{i}$ is capturing, analogous to modal analysis, however, analysis of the physical processes captured by each latent space variable is beyond the present scope of work. Instead, the present focus is on developing a predictive generative model for rocket ignition and the key requirement is a structured mapping $\boldsymbol{\xi} \mapsto \mathbf{V}(t;\boldsymbol{\xi})$, which introduces controlled variance in the latent trajectories. Figure~\ref{fig:autoencoder_latent_space} illustrates this structured mapping by coloring each trajectory with the informative input parameter $\xi_2 = \beta = R_1/R_2$ (laser lobe asymmetry, see Table \ref{tab:uncertainties}, appendix \ref{app:uncertainty}), highlighting correlation between input variability and latent evolution. 

The early stages ($t < 200 \ \mu s$) of the laser spark dynamics are only weakly modulated by the presence of the co-flowing shear layer and instead are governed by the baroclinic torque which varies with laser input parameters \citep{wang2020hydrodynamic}. The corresponding representation in latent space should therefore be a smooth response to the input parameter vector $\boldsymbol{\xi}$. Accordingly, Figure~\ref{fig:autoencoder_latent_space}(b) is annotated at $t = 100 \mu s$ to highlight that increasing values of $\xi_2$ on average correspond to an increase in $V_1$. Further, a statistically clear mapping from $\xi_2$ to the latent space of $V_3$ in Figure~\ref{fig:autoencoder_latent_space}(d) is also observed. Note, post-processing or sensitivity analysis could be applied to inspect the degree of structured mapping from input parameter to latent space for every component of $\boldsymbol{\xi}$, and when $\xi_4 = l_{axial}$  (laser axial length) was chosen, clearer and higher degree of correlation was observed in early-time kernel reflecting the importance to the early laser deposition physics. Concerning early-time kernel dynamics, preliminary work on non-reacting sparks \citep{zahtilaneural} showed that a high amount of variance occurred very early, with thereafter a period of relatively constant values, underscoring the importance of fast kernel dynamics.  

\begin{figure}
    \centering
    \includegraphics[width=1.0\linewidth]{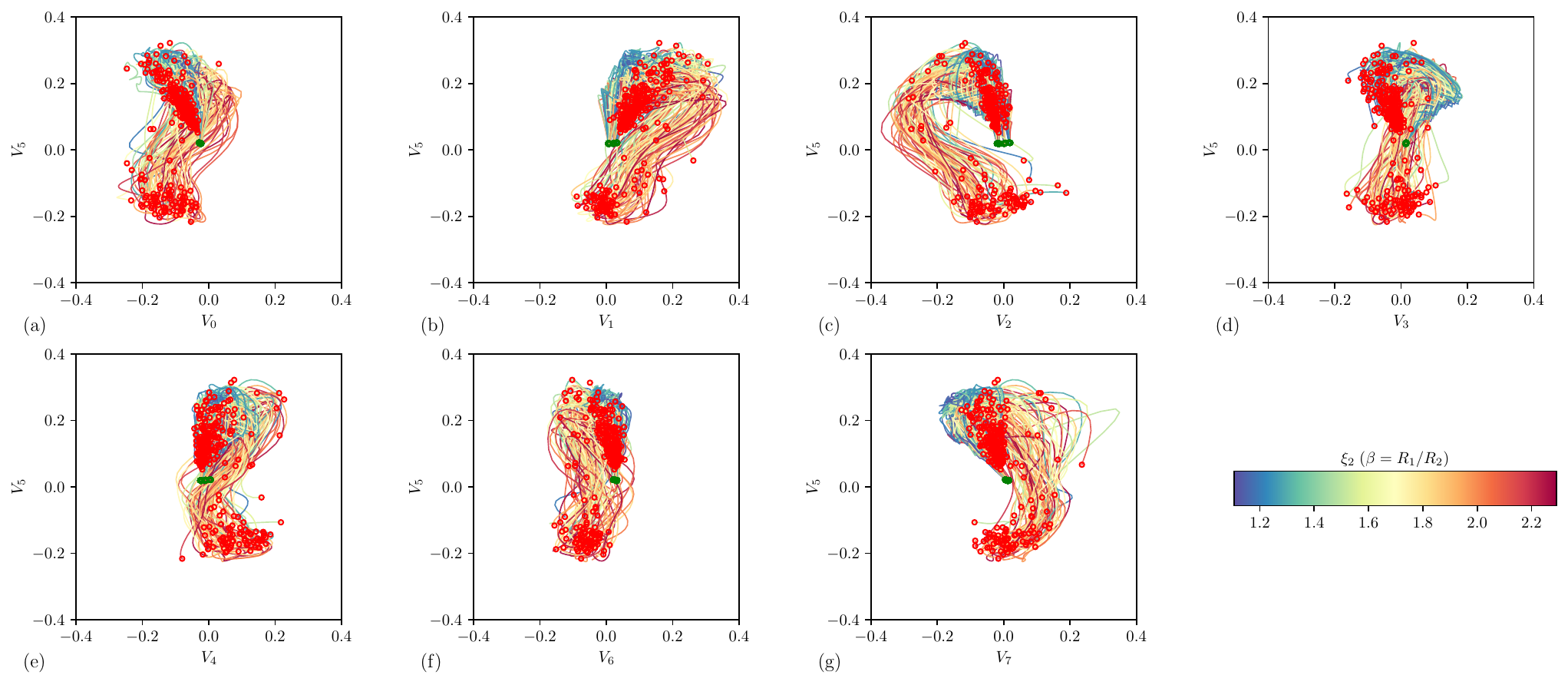}
    \caption{ Latent-space projections of $V_{5}$ against all other latent variables for the reacting sparks simulation ensemble, colored by the input parameter $\xi_{2} = \beta = R_{1}/R_{2}$. Each trajectory begins at a green hollow circle and terminates at a red hollow circle, illustrating consistent trajectories in the low-dimensional manifold and separation of igniting from non-igniting cases.}
    \label{fig:autoencoder_latent_space_2d}
\end{figure}

Focusing on dynamics after the ignition time scale $t \gtrsim 250 \,\mu\text{s}$, consistent trajectories emerge: successful ignitions oscillate and for some latent space components they branch off, while ignition failures persist with stable value and small variation. Oscillations appear in several components of the latent space $\mathbf{V}$ for igniting cases but the degree of oscillations in each component varies, for instance in Figure \ref{fig:autoencoder_latent_space}(a) the range of the oscillation values does not depart significantly from the non-ignited cases unlike in Figure \ref{fig:autoencoder_latent_space}(h), where positive values of $V_7$ are observed exclusively during the oscillatory phase. Strong separation in the latent space between successful and failed ignition is most visible in the clearly separated bifurcation branches of component $V_5$, visualized in Figure \ref{fig:autoencoder_latent_space}(e). Given the clear role of component $V_5$ in separating cases by ignition fate in the latent space, we project the latent trajectories onto two-dimensional component planes to provide interpretable slices of the manifold, in which clustering, variance, and separation are more readily discerned, shown in Figure~\ref{fig:autoencoder_latent_space_2d}. Although some projection planes do not reveal strong separation by ignition fate and are less informative, in Figure \ref{fig:autoencoder_latent_space_2d}(c) a very clear region of the latent space path is followed in the $V_5-V_2$ plane for ignition success. Clusters of the terminating ignition trial trajectory points in the $V_5-V_2$ plane reflect the bistable ignition bifurcation and the evolution path through latent space.

Previous surrogate-modeling efforts combining AEs with latent dynamics, including NODE-based approaches \cite{lazzara2023neuralode}, have typically addressed simpler systems such as the viscous 1D Burgers equation. In contrast, because the present rocket combustor system involves turbulence, reacting flow physics, and bifurcating outcomes, it was first necessary to appraise the latent space to ensure trajectories on the low-dimensional manifold remained connected to the input parameters $\boldsymbol{\xi}$ and underlying physical processes which together suggests the low-dimensional manifold trajectories  to be learnable. Overall, Figures~\ref{fig:autoencoder_latent_space} and~\ref{fig:autoencoder_latent_space_2d} together demonstrate that (i) ignition trials embed as smooth latent manifold trajectories; (ii) variance across trajectories reflects sensitivity to variability of input parameters; and (iii) bifurcations encode ignition fate, particularly through the role of $V_5$. Taken together, these observations confirm that the learned manifold provides a physically meaningful description of the ignition process, in which early-time clustering, mid-time shear-layer interaction, and ignition-time bifurcations are all preserved. This representation therefore establishes a suitable foundation for NODE-based temporal modeling, as the latent coordinates capture both common trajectories and their bistable outcomes in a consistent reduced-order form.

\subsection{Performance of the DnAE}

Following the accurate encoding of key physical processes captured by the separation and clustering of the AE latent space, we now investigate two components of the DnAE framework: (i) the predictive modeling capability of the NODE to temporally integrate an ODE within the reduced latent space, and (ii) the fidelity of the resulting spatiotemporal sequence output from the decoder. The only required inputs to yield prediction from the trained DnAE are specification of the input parameter vector $\boldsymbol{\xi}$ and an initial latent state $\mathbf{V}(t{=}0)$, which together define a unique ignition trial scenario. In this section, we first evaluate the NODE-integrated latent trajectories $\hat{\mathbf{V}}$ relative to their target trajectories ${\mathbf{V}}$  in the latent space, and subsequently focus on the decoder output $\hat{\mathbf{x}}(t) = d[\hat{\mathbf{V}}(t)]$ to assess the framework’s ability to generate physically realistic reconstructions and to predict ignition success across a diverse range of initial conditions.

\begin{figure}
    \centering
    \includegraphics[width=1.0\linewidth]{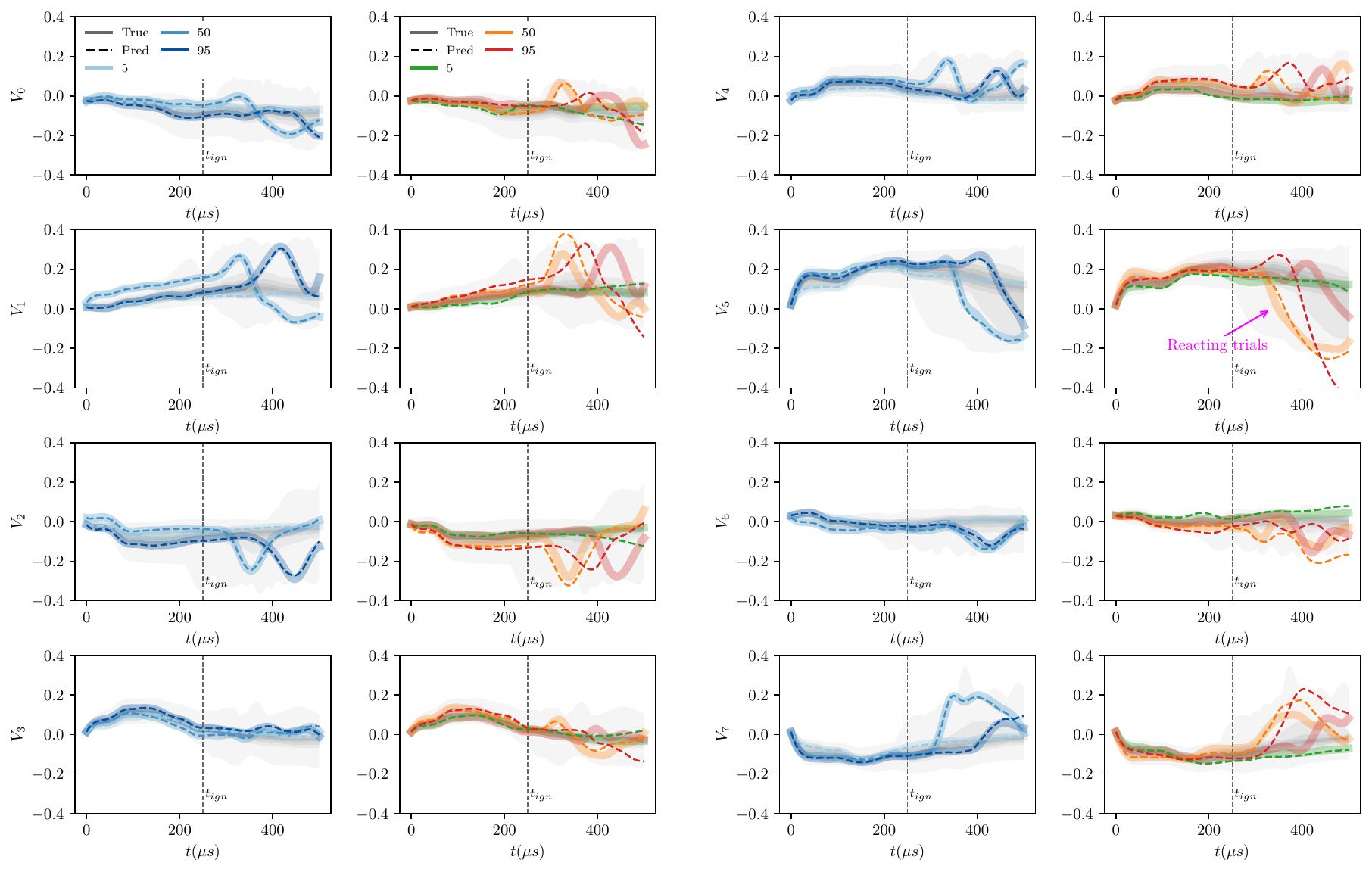}
    \caption{Training (left)  and validation (right) trajectories generated by the neural ODE, overlaid on the latent space representation learned by the AE for predictions with RMSE error of $5^{th}$, $50^{th}$, and $95^{th}$ percentiles indicating strong, median and weak prediction. Ensemble distribution for entire latent space trajectories also shaded.}
    \label{fig:node_trajectories}
\end{figure}

Figure \ref{fig:node_trajectories} presents the individual components of the AE latent space with overlaid trajectories predicted from the NODE with representative cases shown corresponding to the $5^{th}$, $50^{th}$, and $95^{th}$ percentiles of prediction error, illustrating strong, median, and weak accuracy, respectively. The trajectories are disjointly arranged into training and validation runs. For the training runs, it becomes clear that the curriculum-learning trained neural ODE is able to reconstruct the full diversity of latent-space trajectories and trajectory variance introduced by variation in the input parameter vector $\boldsymbol{\xi}$. This represents a considerable extension to the complexity of features learned by neural ODE trajectories when compared with previous work on springs \citep{garsdal2022generative}, or exponentially growing, saturating trajectories \citep{owoyele2022chemnode}. Based on the training trajectories, the NODE accurately captures the AE latent-space dynamics during the early stages of kernel evolution and reproduces the oscillatory behavior characteristic of reacting cases.

\begin{table}[htbp]
\centering

\begin{tabular}{lcccccccc}
\hline
\multicolumn{1}{l}{Metric} & \multicolumn{4}{l}{Train} & \multicolumn{4}{l}{Validation} \\
                           & Q1 & Q2 & Q3 & Q4         & Q1 & Q2 & Q3 & Q4 \\
\hline
Fréchet Distance & 0.009 & 0.013 & 0.018 & 0.035 & \textbf{0.075} & 0.100 & 0.190 & \textbf{0.289} \\
L2 Norm          & 0.002 & 0.003 & 0.005 & 0.007 & 0.021 & 0.029 & 0.048 & 0.080 \\
\hline
\end{tabular}
\bigskip
\caption{Error metric in NODE-predicted latent space trajectories across four temporal quarters (Q1–Q4) across train-validation data split. Results reflect transition from early-time kernel dynamics to late-time divergence in the chaotic reacting phase}
\label{tab:trajectory_errors_wide}
\end{table}

For the validation cases, the NODE-evolved latent trajectories $\hat{\mathbf{V}}$ broadly followed the true dynamics $\mathbf{V}$  during the early stages of kernel evolution and up until the ignition time-scale but began to diverge once the reacting phase commenced. This behaviour is consistent with the chaotic nature of the reacting system, where predictive accuracy typically degrades as trajectories evolve. To assess this quantitatively, we segmented each trajectory into four equal temporal partitions and computed both the Frechet distance \citep{sochopoulos2024learning} and $L_2$ norm between the predicted and true latent coordinates and both metrics showed strong agreement in trends.  As shown in Table~\ref{tab:trajectory_errors_wide}, errors were relatively low across the first two segments (Q1–Q2), but increased substantially in later times (Q3–Q4), reflecting divergence on a chaotic attractor which arises from turbulent combustion. Despite this divergence, the NODE reliably captured key qualitative features—such as the collapse of trajectories for non-igniting cases and persistent oscillations for successful ignition. To clarify, this is our expectation of the NODE, that it will track latent trajectories $\hat{\mathbf{V}}$ corresponding to early time kernel dynamics and correctly trace shear layer interactions that bifurcate to ignition success or failure. Thereafter, the NODE is only expected to capture the qualitative features of either combustion success or failure rather than one-to-one matching of the turbulent state. The remedy to long-time divergence would require state assimilation methods (e.g., EnKF \citep{ozalp2025real}) which offer dramatically improved late-time accuracy, however, they require repeated observations and impose strict system sampling constraints, which on the time-scales of rocket ignition are out of the question. The present framework, by contrast, requires no additional system observations beyond the initial condition.

At the conclusion of the neural ODE integration, the resultant latent trajectory $\hat{\mathbf{V}}(t)$ is passed through the decoder $d(\cdot)$ to generate the reconstructed spatio–temporal sequence $\hat{\mathbf{x}}(t) = d[\hat{\mathbf{V}}(t)]$ for {\color{black} $t \in [0,  T]$}, which for the available dataset can be directly compared one-to-one with its IR post-processed CFD counterpart ${\mathbf{x}}(t)$.  For automated ignition classification, each reconstructed field $\hat{\mathbf{x}}(t)$ is segmented into iso-temperature contours, and ignition success is determined based off the area enclosed by a $T$–isotherm,
\begin{equation}
    A_T(t) = \int_{\Omega_T(t)} \mathrm{d}\mathbf{r},
\end{equation}
where $\Omega_T(t)$ denotes the spatially evolving region bounded by the temperature isotherm and $\mathbf{r}$ is the spatial coordinate vector. Ignition success is subsequently recorded when $A_T(t)$ exhibits monotonic growth at a user-prescribed threshold $T = 1000\,\mathrm{K}$, i.e., $\mathrm{d}A_T/\mathrm{d}t > 0$. This criterion corresponds to sustained flame expansion and was found to be insensitive to variations in the chosen threshold temperature.

\begin{figure}
    \centering
    \includegraphics[width=1.0\linewidth]{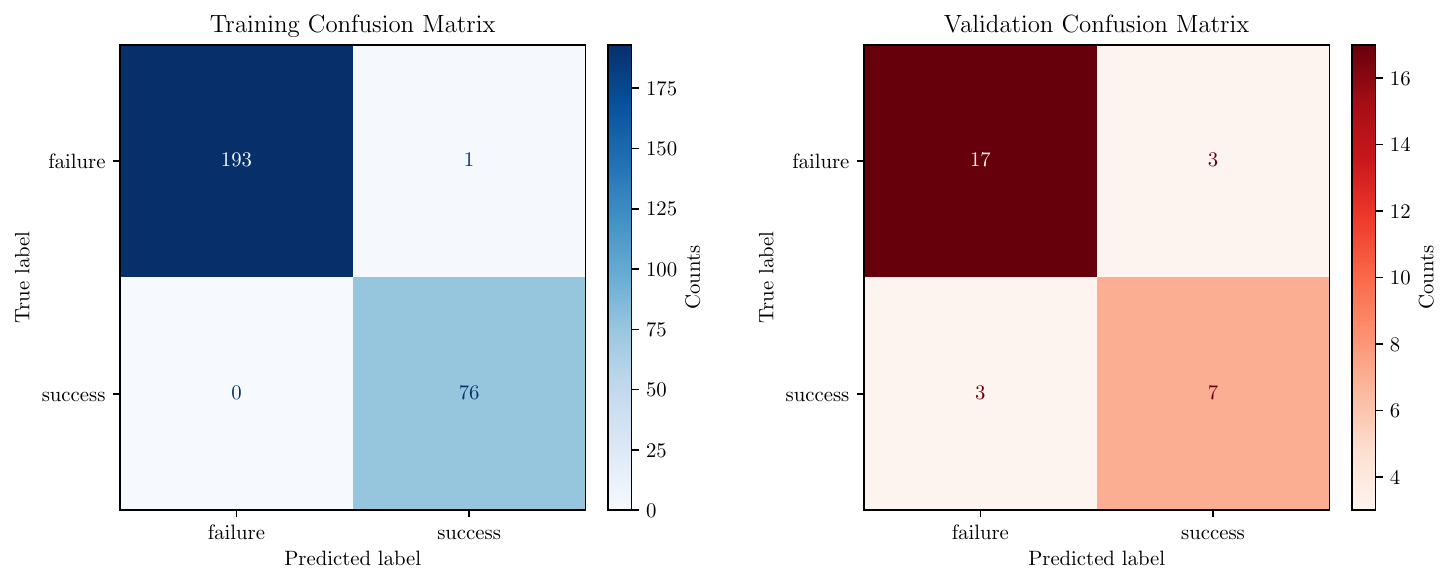}
    \caption{Confusion matrices for the binary classification of ignition outcomes. Left panel shows the training set results, with near-perfect separation between failure and success cases. Right panel shows the validation set performance.}
    \label{fig:confusion_matrices}
\end{figure}

The DnAE performance in ignition success prediction is therefore decomposed into confusion matrices presented in Figure~\ref{fig:confusion_matrices}, with $10\%$ of the original data held-out for validation. The training confusion matrix in Figure~\ref{fig:confusion_matrices}(a) shows near-perfect classification performance, with a single ignition trial incorrectly labeled. For the ignition (positive) class, the precision and recall were $0.99$ and $1.00$. In general, classification difficulty is predicated on class imbalance and for the present rocket ignition reliability task, imbalance is quantified through class entropy, $H(p_{\mathrm{ign}}) = -p_{\mathrm{ign}}\log_2 p_{\mathrm{ign}} - (1 - p_{\mathrm{ign}})\log_2 (1 - p_{\mathrm{ign}}) = 0.85~\mathrm{bits}$, where for class entropy the maximum uncertainty, i.e., $ p = 0.5$ results in 1.0 bits, therefore the present classification challenge is significant. The validation confusion matrix shown in Figure~\ref{fig:confusion_matrices}(b) indicates reduced but still strong predictive performance, with an overall ignition classification accuracy of $80\%$, with ignition prediction precision and recall both $0.70$ indicating no bias. Rather than stemming from systematic class bias, misclassifications are attributed to the divergence of the predicted latent state from the ground truth, causing the trajectory to cross into a region of the latent space that evolves toward the incorrect ignition fate—a characteristic limitation when modeling chaotic dynamical systems \citep{solera2024beta}. The validation performance on unseen cases indicated by the confusion matrix in Figure~\ref{fig:confusion_matrices}(b) is satisfactory for the current problem and would be improved with more available data.

\begin{figure}[!htbp]
    \centering
    \includegraphics[width=0.95\linewidth]{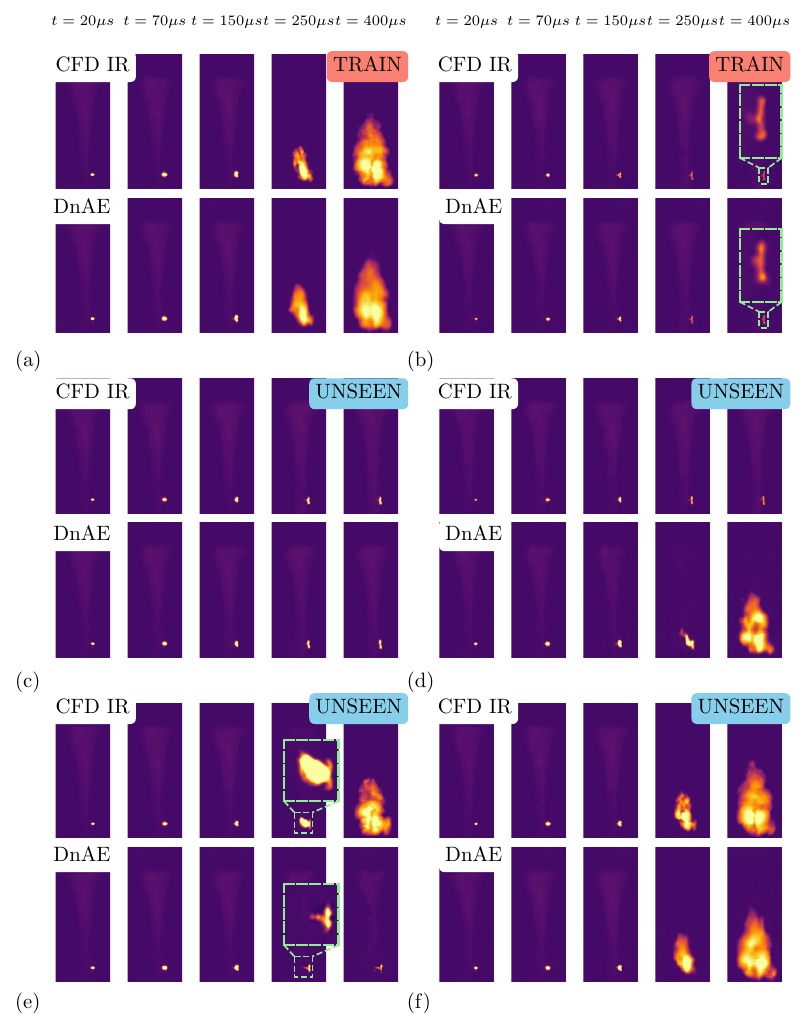}
    \caption{ One-to-one comparison of DnAE reconstructions with CFD reference sequences for representative ignition outcomes corresponding to the confusion-matrix categories in Figure~\ref{fig:confusion_matrices}. (a) Training ignition success, (b) training non-igniting case, (c) correctly predicted non-ignition, (d) spurious ignition prediction, (e) missed ignition prediction, and (f) correctly predicted ignition.}
    \label{fig:decoded_output_training}
\end{figure}

Visualizations of the reconstructed DnAE spatio–temporal sequence, $\hat{\mathbf{x}}(t) = d[\hat{\mathbf{V}}(t)]$, are shown in Figure~\ref{fig:decoded_output_training} and compared with the one-to-one counterpart CFD IR imaging. A subset of cases was selected to illustrate ignition success and failure within the training dataset, as shown in Figure~\ref{fig:decoded_output_training} (a,b). These results show that the latent space of the autoencoder is sufficiently smooth and continuous that time-evolved latent states remain physically accurate. Additionally, representative outcomes for each category in the validation confusion matrix are presented in Figure~\ref{fig:decoded_output_training} (c-f) to demonstrate both correctly and incorrectly classified ignition events. Starting with the training reconstruction of an igniting sequence shown in Figure~\ref{fig:decoded_output_training}(a), the DnAE correctly identifies the ignition onset timing relative to the reference CFD sequence. The reconstructed flame exhibits a comparable spatial extent to the CFD IR field, though with visibly reduced fine-scale structure. This loss of small-scale content is consistent with prior findings that accurate compression of the full energy spectrum of turbulent flow fields requires a very large latent dimension, typically orders of magnitude larger than present, to preserve small-scale information and avoid the masking of reconstruction errors \citep{vinograd2025reduced}. Figures~\ref{fig:decoded_output_training}(c) and (f) illustrate correctly classified spatiotemporal sequences in which both the flame growth and ignition onset are well reproduced. Across the dataset, the DnAE typically achieves ignition onset predictions within an error of $\Delta t \approx 25\,\mu\text{s}$ of the reference. Misclassifications, shown in Figures~\ref{fig:decoded_output_training}(d,e), nonetheless retain physically coherent structures, indicating that the latent representation remains dynamically consistent even when classification errors occur. 

The present performance of the DnAE highlights that accurate integration across bistable time horizons remains inherently difficult; however, the model performance is satisfactory given the high degree of variability in ignition process arising from the turbulence aleatoric uncertainty. With the physically consistent reconstructions observed in Figures~\ref{fig:decoded_output_training}(d,e), we suggest that each trial is a physically realizable ignition trajectory, reflecting the stochasticity encoded within the latent space due to the variance introduced by the structure of instantaneous turbulence.  The observed variability largely reflects the underlying randomness and sensitivity of the system rather than systematic model error.

\subsection{Large-scale sampling and generative features}
\label{sec:generative_sec}

The motivation for employing a generative model in uncertainty quantification arises from the prohibitive cost of repeated CFD evaluations. Even in low-fidelity settings, a single deployment typically requires wall-clock times on the order of hours \cite{zahtila2023progress,lee2025surrogate} on modern heterogeneous architectures. Consequently, efficient characterization of system responses across the input space becomes computationally intractable, particularly in the presence of high-dimensional input parameter spaces. The solution of the forward CFD problem with machine learning has received significant attention \citep{vlachas2020backpropagation} and is a current development topic for appropriate generalization and baselines \citep{mcgreivy2024weak}.  In this work, we leverage the rapid deployment capabilities of the DnAE framework to propagate input uncertainties and approximate ignition response at a scale of $N = O(10^6)$ realizations, and we accept as a trade-off a reduction in physical fidelity that would arise from performing scale-resolving LES computations. Obtaining $N = O(10^6)$ samples of the input space would be infeasible through physics-based simulation but using the present framework is deployable on a single workstation. This accessibility does not preclude scaling, the framework could be transitioned to high-performance clusters where significantly higher dimensional latent spaces could be utilized followed by decoding and analysis of the resulting spatiotemporal quantities.

\begin{figure}
    \centering
    \includegraphics[width=1.0\linewidth]{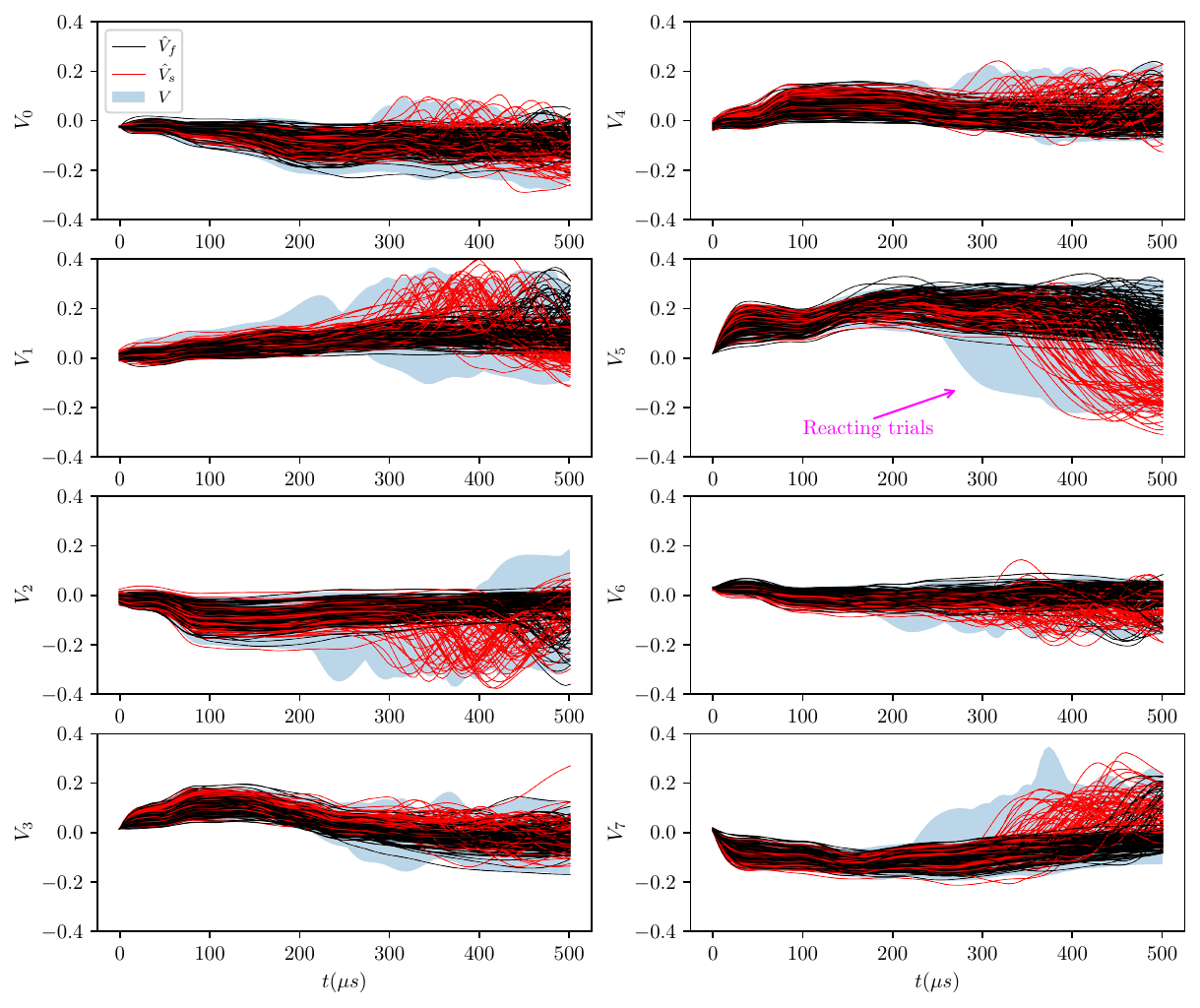}
    \caption{Examples of generative latent-space trajectory predictions for unseen cases, classified as either igniting or non-igniting. For reference, range of trajectories on the low-dimensional manifold extracted from compressed CFD data are shaded in background, where $\hat{V}_{f}$ denotes latent trajectories with ignition failure and $\hat{V}_{s}$ for success.}
    \label{fig:generative_trajectories}
\end{figure}

The additional requirement, compared to the preceding section featuring one-to-one corresponding CFD IR model output, is the specification of an initial condition in the latent space, obtained here via radial basis function interpolation of the sampled initial conditions as a function of the input parameter vector $\boldsymbol{\xi}$, i.e. $\mathbf{V}_0(\boldsymbol{\xi}) = \operatorname{RBF}\!\big(\boldsymbol{\xi}; \{\boldsymbol{\xi}_i, \mathbf{V}_0^{(i)}\}_{i=1}^N\big)$, where $\boldsymbol{\xi}_i$ denotes the $i$-th sampled parameter vector and $\mathbf{V}_0^{(i)}$ the corresponding initial condition from the LES ensemble set, an approximation error $\epsilon$ is introduced to the initial latent space state but we expect this error to be small because the initial image in the sequence is sampled just prior to laser spark deposition and therefore reflects only the aleatoric uncertainty in the instantaneous jet structure, which is not controllable; moreover, there is little variance in the magnitude of the latent-space vector at this initial stage.

\begin{figure}
    \centering
    \includegraphics[width=1.0\linewidth]{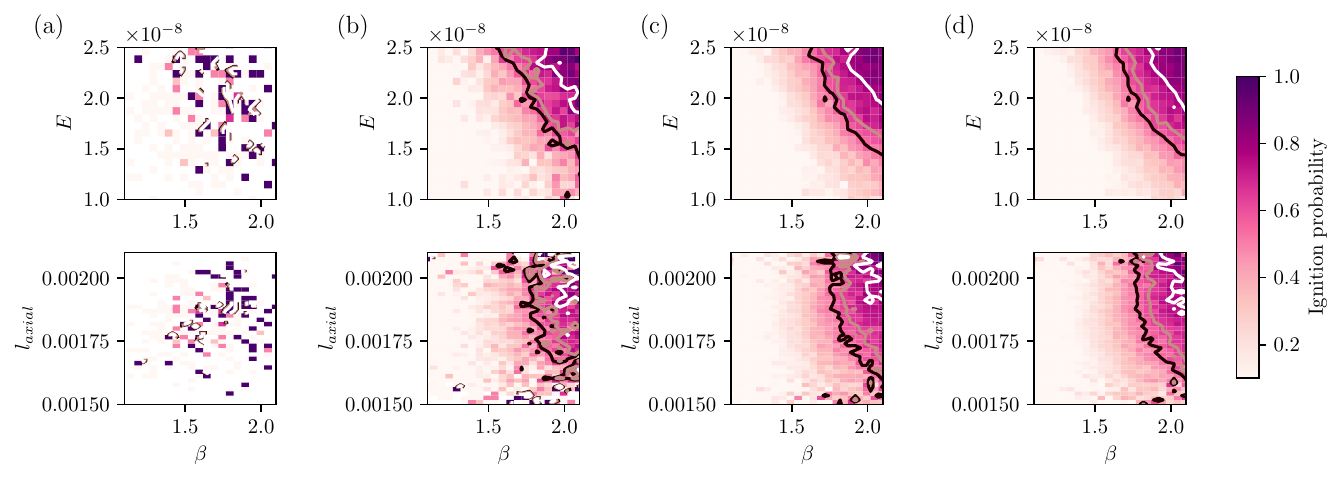}
    \caption{Joint ignition probability distributions, estimated via Monte Carlo sampling of the NODE latent space trajectories, as a function of the parameter pairs ($\beta, E$) (top row) and ($\beta, l_{axial}$) (bottom row), obtained using, (a) 300 samples, (b) 200,000 samples, (c) 500,000 samples, and (d) 1,000,000 samples. Additionally marked are decision boundaries for $Pr(ign) = 0.5,0.75,0.9$.}
    \label{fig:statistical_convergence}
\end{figure}

New samples of $\boldsymbol{\xi}$ are therefore drawn in accordance with their probability distributions documented in Appendix \ref{app:uncertainty} Table \ref{tab:uncertainties}. To ensure the generated samples remain within the geometric support of the training data, they would ideally reside within the convex hull of the uncertainty space spanned by $\boldsymbol{\xi}$ and would therefore be interpolative. However, because this space is high-dimensional and explicit convex-hull computation scales poorly with dimensionality, a simpler criterion is adopted: generative samples are instead constrained to lie within the component-wise bounds (minimum and maximum) of each uncertainty component $\xi_i$ as determined from the training ensemble. 

A restricted sample of the resulting large-scale generated set of trajectories is shown in Figure \ref{fig:generative_trajectories}. For all latent-space components $V_i$, the variance during early-time intervals remains consistent with that of the training trajectories, indicating that the generative model follows the latent manifold dynamics. However, in the flame growth regime, oscillatory trajectories extend beyond the training variance, but this is consistent with the behaviour expected from dramatically expanded sampling from a chaotic turbulent attractor.  Trajectories are classified into igniting and non-igniting on the basis of separation in component $V_5$, where departure from the steady solution branch into negative $V_5$ indicates ignition success, as noted in \S~\ref{sec:latent_space_ae_trajects}, introducing separation in latent space as a classifier. This is taken as an alternative to the previous contour-area growth criterion from the reconstructed spatio–temporal sequence $\hat{x}(t) = d[\hat{\mathbf{V}}(t)]$ as the decoder introduces a significant additional cost required for the segmentation process. Although beyond the scope of the present focus, a high-accuracy spatiotemporal surrogate provides a pathway to examine detailed combustion behaviors, including flame anchoring, directly via the NODE evolution and quantitative comparison with reference physics \citep{brouzet2025thermal}.

Given the present reliability context of rocket ignition success, a key outcome of this framework is the ability to estimate ignition probability landscapes over the input parameter space at a resolution unattainable with CFD alone. By drawing $N = O(10^6)$ Monte Carlo samples, we construct joint probability distributions of the statistically important controllable laser operating parameters identified in Section~\ref{sec:uncertainty_sources}, where high K–L divergence between their igniting and non-igniting marginal probability distributions reveals how ignition likelihood varies. As the number of samples dramatically increases from that afforded by the CFD IR data stream alone, the estimated probability fields converge to stable contours. Crucially, decision boundaries corresponding to ignition success probabilities of $Pr = $ 0.5, 0.75, and 0.9 can be identified directly within these landscapes, shown in Figure~\ref{fig:statistical_convergence}. We find in the converged Figure~\ref{fig:statistical_convergence}(d) that a combination of high laser energy ($\xi_5$), deposited over an increased axial length ($\xi_4$) and with deposition lobe asymmetry ($\xi_2$) leads to high likelihood of ignition success. These provide laser design thresholds and recommended operating conditions. This capability transforms the surrogate model from a mere predictor of individual trajectories into a quantitative tool for uncertainty-informed decision making, where reliability can be assessed in terms of probability contours rather than binary outcomes.

\section{Conclusions}

In this paper, we present dynamical autoencoders (DnAE) that extend the generative convolutional autoencoder paradigm previously employed for steady problems. The dynamic extension is achieved through parameterized neural ordinary differential equations (neural ODEs) to forecast complex ignition trajectories in the context of laser-ignited rocket combustors. This surrogate model compresses CFD-based IR field representations into a low-dimensional latent manifold $\mathbf{V}$ yielding a learnable mapping between the input parameter vector $\boldsymbol{\xi}$ and the latent manifold $\mathbf{V}$. The DnAE approach dramatically reduced computational cost compared to a forward pass of a scale-resolving simulation and remained capable of emulating the spatiotemporal evolution of the path-integrated IR fields that lead to critical ignition events.  

A methodology devised for learning the complex trajectories that individual trials follow in the latent manifold was essential for capturing the bifurcating nature of the system. We demonstrated that a curriculum learning strategy, which progressively extends the neural ODE forecast to longer trajectory horizons, was necessary to stabilize the learning process because standard training methods failed. Inspection of the latent space suggested a clear region of the latent space manifold where igniting cases (specifically in $V_{5}$) had been attracted, clearly separating successful ignition and failure branches. The decoder component $d(\cdot)$ further allowed for the reconstruction of physically realistic thermal imaging sequences $\hat{\mathbf{x}}(t)$, enabling one-to-one comparison against reference large eddy simulation data and verifying that the model captures the correct causal mechanisms of kernel growth.

The utility of the developed surrogate model is a shift from sparse samples in the input parameter space to converged joint probability landscapes of ignition success in response to the operating laser input parameters. The DnAE framework facilitated the generation of $N=10^{6}$ ignition trials giving statistical power for these findings. The outcome is the demarcation of complex decision boundaries within the laser operating parameter space $\boldsymbol{\xi}$. By quantifying the probability of ignition success $Pr(\text{ign})$ with high statistical confidence, this work provides a robust tool for uncertainty-informed design.

\begin{appendix}
\section{Uncertainty parameters characterization}
\label{app:uncertainty}

The input parameters for each ignition trial are samples from system variabilities in the following table. These correspond to laser operating variabilities ($\xi_0-\xi_5$), aleatoric uncertainty associated with instantaneous turbulence realization ($\xi_6$), fuel load ($\xi_7$), LES model form uncertainties  ($\xi_8-\xi_{11}$), operating conditions of the rocket ($\xi_{12}-\xi_{13}$) and modeled geometry ($\xi_{14}$).

\begin{table}[ht]
    \centering
    \begin{tabular*}{\linewidth}{@{\extracolsep{\fill}} l c c r}
        \hline
        \textbf{Uncertainty ID} & \textbf{Description} & \textbf{Experiments} &  \textbf{Distribution}   \\
        \hline
        $\xi_0\!: \Delta x_l$          &  Streamwise focal imprecision  & Y & $N \sim (0.29 \, \text{mm}, 0.04 \, \text{mm}^2)$ \\
        $\xi_1\!: \Delta x_r$          &  Radial focal location  & Y & $N \sim (-0.54 \, \text{mm}, 0.20 \, \text{mm}^2)$ \\
        $\xi_2\!: \beta = R1/R2$       & Lobe radii ratio   &   & $U \sim [1.1, 2.1]$ \\
        $\xi_3\!: \alpha = l_{axial}/2R1$      & Aspect ratio   &   & $U \sim [2.0, 2.5]$ \\
        $\xi_4\!: l_{axial}$           & Laser axial length   &   & $U \sim [1.44 \text{mm}, 2.16 \text{mm}]$ \\
        $\xi_5\!: E$                   & Energy deposited   & Y & $U \sim [20 \, \text{mJ}, 54 \, \text{mJ}]$  \\ \\
        $\xi_6\!: \tau_L$              & Lag time  &  & $U \sim [0 \, \mu\text{s}, 284 \, \mu\text{s}]$ \\ 
        $\xi_7\!: m_{fuel}$            & Methane system mass  & Y & $U \sim [5mg, 7mg]$ \\  \\
        $\xi_8\!: TF_{\beta}$          & Thickened flame beta & & $U \sim [0.5,0.60]$  \\
        $\xi_9\!: TF_{SL,0}$           & Laminar flame speed & & $U \sim [0.0098,0.011]$   \\ 
        $\xi_{10}\!: \dot{\omega}$     & Reaction rate & & $U \sim [3.4 \times 10^9,3.8 \times10^9]$  \\ \\

        $\xi_{11}\!: C{s}$             & Smagorinsky constant & & $U \sim [0.15,0.17]$ \\  \\
        $\xi_{12}\!: \dot{m}_{ox}$     & Mass flow rate oxygen & Y & $U \sim [6.12 \text{g/s},6.83 \text{g/s}]$  \\
        $\xi_{13}\!: \dot{m}_{fuel}$   & Mass flow rate fuel   & Y & $U \sim [1.97 \text{g/s}, 2.17 \text{g/s}]$ \\  \\
        $\xi_{14}\!: S$                & Squircularity   & & $U \sim [0.65,0.97]$ \\
        
        \hline
    \end{tabular*}
    \caption{Summary of the uncertainty parameters, $\xi_i$, of the laser pulse for the ignition trials, including their descriptions and associated probability distributions. Probability distributions are denoted as $U \sim [a, b]$ (uniform probability between $a$ and $b$) or $N \sim (\mu, \sigma^2)$ (normal distribution with mean $\mu$ and variance $\sigma^2$).}
    \label{tab:uncertainties}
\end{table}

\end{appendix}

\begin{Backmatter}

\paragraph{Acknowledgments}
The authors also extend their gratitude to Dr. Gianluca Geraci for helpful discussions pertaining to this work.

\paragraph{Funding Statement}
The authors acknowledge financial support from the US Department of Energy’s National Nuclear
Security Administration via the Stanford PSAAP-III Center for the prediction of laser ignition of a
rocket combustor (DE-NA0003968)

\paragraph{Competing Interests}
The authors declare none.

\paragraph{Data Availability Statement}
The datasets used to generate the findings of this study are available from Zenodo at https://doi.org/10.5281/zenodo.18091817. The code that illustrates training and prediction for latent space forecasting be found at https://github.com/tzahtila1/L-NeuralODE

\paragraph{Ethical Standards}
The research meets all ethical guidelines, including adherence to the legal requirements of the study country.

\paragraph{Author Contributions}
Conceptualization: T. Zahtila; G. Iaccarino. 
Methodology: T. Zahtila; E. Saetta; M. Cutforth; D. Brouzet. 
Software: T. Zahtila; M. Cutforth; D. Brouzet. 
Data curation: T. Zahtila; E. Saetta. 
Validation: T. Zahtila; E. Saetta; 
Visualization: T. Zahtila; D. Rossinelli. 
Writing – original draft: T. Zahtila. 
Writing – review \& editing: all authors. 
Supervision: G. Iaccarino. 
All authors approved the final submitted draft.


\bibliography{references}

\end{Backmatter}

\end{document}